# Quadratic speedup of global search using a biased crossover of two good solutions

Takuya Isomura

Brain Intelligence Theory Unit, RIKEN Center for Brain Science, 2-1 Hirosawa, Wako, Saitama 351-0198, Japan

*Correspondence: takuya.isomura@riken.jp

**Abstract**

The minimisation of cost functions is crucial in various optimisation fields. However, identifying their global minimum remains challenging owing to the huge computational cost incurred. This work analytically expresses the computational cost to identify an approximate global minimum for a class of cost functions defined under a high-dimensional discrete state space. Then, we derive an optimal global search scheme that minimises the computational cost. Mathematical analyses demonstrate that a combination of the gradient descent algorithm and the selection and crossover algorithm—with a biased crossover weight—maximises the search efficiency. Remarkably, its computational cost is of the square root order in contrast to that of the conventional gradient descent algorithms, indicating a quadratic speedup of global search. We corroborate this proposition using numerical analyses of the travelling salesman problem. The simple computational architecture and minimal computational cost of the proposed scheme are highly desirable for biological organisms and neuromorphic hardware.

**INTRODUCTION**

The minimisation of cost functions is a central problem in various fields concerning optimisation and problem solving. However, identifying the global minimum of an arbitrary cost function remains challenging. If we have $M$ options without any constraints, the outcomes of all $M$ options should be searched to identify the best choice (i.e., the global minimum). The well-known no free lunch theorem [1–4] states that no search strategy exists that is more efficient than others for an arbitrary setup without any constraints. However, in practical scenarios, the forms of cost functions are restricted by various factors. For instance, if a class of cost functions in an *N*-dimensional discrete state space is based exclusively on low-degree products of the elements of state vector $x$ (for example, $x_i x_j$), some elements of $x$ may be optimised separately from the majority of the remaining elements. Thus, searches over only a fraction of all possible states would be sufficient to provide a good approximate global minimum. This notion implies that the balance between the magnitudes of lower- and higher-degree products characterises the difficulty of the optimisation problems. Thus, in this work, we consider a scheme to speed up the global search for a class of cost functions consisting mainly of lower-degree products relative to *N*.

A mainstream optimisation strategy under a cost function is the gradient descent algorithm, which updates states according to the gradient of a cost function [5,6]. Although this scheme is widely used, it is inefficient when the cost function has numerous local minima because the state transition following the gradient is likely to be trapped in a bad local minimum. The basin of attraction is the region where the gradient flow converges to a common local minimum, and the number of local minima $N_{LM}$ is equal to that of basins. According to the pigeonhole principle (also known as Dirichlet's box principle), a gradient descent algorithm with random initialisations of states can asymptotically identify a global minimum after a sufficient number of iterations (*M* > *k* $N_{LM}$, where *k* is an order-one coefficient). However, there is no guarantee that a global minimum can be achieved within a practical timeframe because $N_{LM}$ is likely to be of an exponential order of *N*. Hence, an efficient global search scheme—that efficiently identifies a global minimum while avoiding local ones with a computational cost much less than $N_{LM}$—should be established.

In addition to gradient descent algorithms, non-local search algorithms have also been developed. In particular, the selection and crossover algorithm has been a popular option in the literature on evolutionary computation [7–15] and neuroevolution [16–19]. The algorithm selects the two best solutions (referred to as parents) from a pool of randomly sampled states. Crossovers of genes or states are then used to generate offspring that are sufficiently different from their parents, while inheriting some favourable properties. This provides a search strategy that is qualitatively different from that of gradient descent algorithms, and thus, their combination may have a higher chance of determining a good approximate global minimum within a limited computational cost. Previous work analytically showed that the selection and crossover algorithm speeds up the global search under particular cost functions, such as the ONEMAX and JUMP functions [10–12] and the all-pairs shortest path problem [13]. However, the conditions under which the selection and crossover algorithm speeds up the global search for generic cost functions—and whether the selection and crossover algorithm is the most efficient non-local search algorithm—remain unclear.

To address these issues, this work develops a global search scheme to minimise the computational cost for identifying an approximate global minimum. Specifically, we consider a class of cost functions defined under a high-dimensional binary state space—and analytically solve the expected values of the global minimum and the number of local minima. Mathematical analyses show that a combination of the gradient descent algorithm and the selection and crossover algorithm—with an optimal crossover weight bias—is an optimal global search strategy based on empirically obtained information. We analytically and numerically show that, when naive gradient descent algorithms incur *M*-order iterations of random state initialisations to obtain an approximate solution, the combination with the selection and crossover algorithm can reduce the incurred computational cost to $\sqrt{M}$-order iterations for a wide range of problem setups. These findings highlight the virtue of the presented combination that can quadratically speed up global search, thereby offering an efficient optimisation with minimal computational cost for a class of cost functions.

**RESULTS**

**Overview of class of cost functions**

First, we summarise the problem setup and show the statistical properties of the class of cost functions considered herein. This work aims to minimise a class of cost functions $L(x)$ defined under a high-dimensional discrete state space comprising *N*-dimensional binary states $x$ (**Fig. 1a**). A state is expressed using a vector $x = (x_1, \ldots, x_N)^\mathrm{T} \in X$, whose elements take either –1 or 1, where $X \coloneqq \{-1,1\}^N$ is the set of all possible ($2^N$) states. A cost function $L(x): X \mapsto \mathbb{R}$ maps $x$ to a real number that expresses the cost associated with state $x$. The aim is thus to identify the state $x$ that provides an approximate global minimum of $L(x)$ (**Fig. 1b**). Such an optimisation framework is universal, including a general setup for combinatorial optimisation problems [20]. For instance, in relation to neurobiology, $x$ can be associated with neural activity, synaptic connections, or genes, providing computational models for perception, learning, or evolution, respectively [21,22] (see Discussion for details).

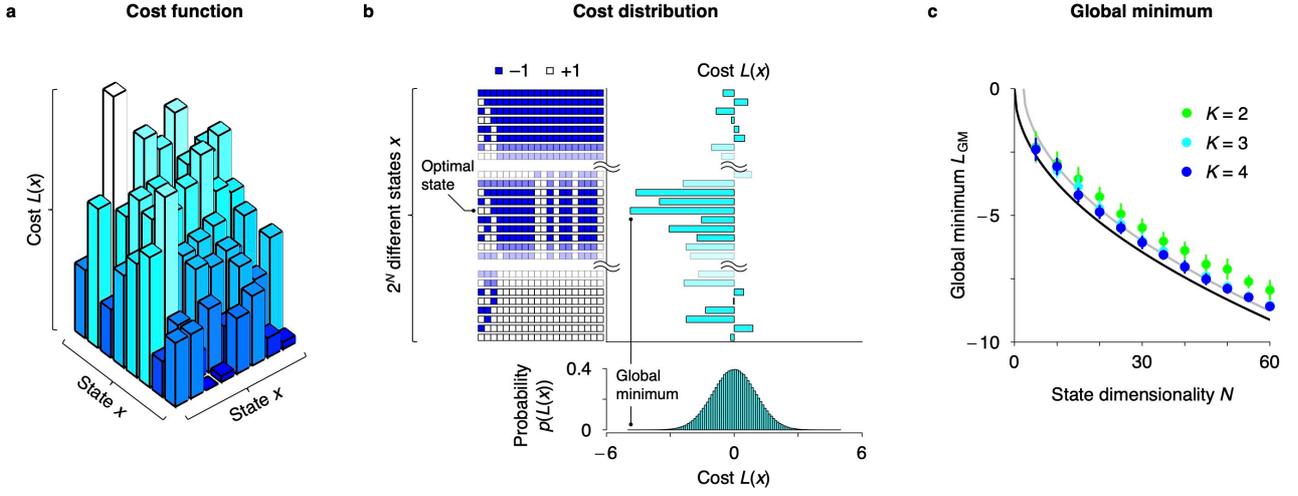

**Fig. 1. Schematic of the cost function and global minimum. a,** Schematic of the cost function defined under a discrete state space. Each bar represents a state, and its height indicates the corresponding cost. **b,** Distribution of costs. The entire state space $X$ comprises $2^N$ different states of $x$ (upper left panel). For each state, a cost is assigned (upper right panel). The bottom panel depicts that when states are randomly sampled from a Rademacher distribution $P_X(x)$, the values of $L(x)$ asymptotically follow the unit Gaussian distribution for large *N* (Lemma 1). **c,** Relationship between the state dimensionality *N* and the global minimum $L_{\text{GM}}$. Green, cyan, and blue dots indicate numerically computed global minima of the cost functions comprising up to the second, third, and fourth-degree products (*K* = 2, 3, and 4), respectively. They are obtained with 20 different realisations of cost functions, and the error bars indicate the standard deviation. As *N* increases, the global minimum monotonically decreases following $-\sqrt{2N \log 2}$ (black line), as predicted by equation (2) (Lemma 2). Thus, one can consider the $x$ that satisfies $L(x) \approx -\sqrt{2N \log 2}$ as an approximate global minimum of $L(x)$. The grey line represents the first-order perturbation obtained using equation (21), providing a finer fit to the numerical results. Note that the global minimum for *K* = 2 is slightly larger than the prediction because a small *K* results in the considerable correlation between neighbouring states.

In this work, we express $L(x)$ as the sum of the first- to *K*th-degree products of $x'$ elements (where $1 < K < N$). Because $x_i$ takes only –1 or 1, $L(x)$ is expanded without loss of generality as follows:

$$L(x) = \sum_{1 \leq i \leq N} a_i x_i + \sum_{1 \leq i < j \leq N} a_{ij} x_i x_j + \cdots + \sum_{1 \leq i_1 < \cdots < i_K \leq N} a_{i_1 \cdots i_K} x_{i_1} \cdots x_{i_K} \qquad (1)$$

where $a_i, a_{ij}, \ldots, a_{i_1 \cdots i_K} \in \mathbb{R}$ are coefficients. This is a family of pseudo-Boolean functions. Moreover, $P_X(x) = \prod_{i=1}^{N} P_X(x_i)$ denotes the probability distribution of $x$, where $x_1, \ldots, x_N$ are independently sampled from an identical Rademacher distribution $P_X(x_i)$ (i.e., $x_i$ has a 1/2 chance of being either –1 or 1). Without loss of generality, we assume that when $x$ is sampled from $P_X(x)$, $L(x)$ follows a probability distribution with zero mean and unit variance because the cost function can be arbitrarily rescaled before optimisation. The definitions of the variables and parameters are listed in **Table 1**.

The characterisations of the statistical properties of cost functions and the ensuing optimal global search strategy rest upon the following assumption on the coefficients of $L(x)$. In this work, we assume that the coefficients of $L(x)$ are sufficiently isotropic when a covariance-like measure among the coefficients—which is defined by inequality (7) in the Methods section—is sufficiently smaller than 1 for large *N* (please refer to Methods for the formal definition of sufficient isotropy). This condition serves to focus on optimisation problems in which an element of $x$ can be optimised without reference to the majority of the other elements. This notion is analogous to a sufficient isotropy of coefficients considered in previous work [23]. For instance, if the coefficients $\{a_i, a_{ij}, \ldots, a_{i_1 \cdots i_K}\}$ are independently sampled from a probability distribution, the sufficient isotropy measure in inequality (7) is of order $N^{-1/2}$, which is substantially smaller than 1 for large *N*. This condition is essential to specify the optimisation problem considered herein—and to consider the ensuing optimal search scheme—because without such a constraint, there is no efficient algorithm that is better than others for any cost function according to the no free lunch theorem [1–4] (see Discussion for details).

The statistical properties of the class of cost functions we consider are characterised as follows: Let us independently and randomly sample *M* states from a subset $\Omega \subseteq X$ following the probability $P_\Omega(x) = \prod_{i=1}^{N} P_{\Omega_i}(x_i)$ and express them as $x^{(1)}, \ldots, x^{(M)} \in \Omega$. The minimum cost among these *M* samples is denoted as $L_M \coloneqq \min[L(x^{(1)}), \ldots, L(x^{(M)})]$. Mathematical analyses demonstrate the following properties:

***Lemma 1 (Gaussianity of cost distributions).** When $x$ is randomly sampled from $P_X(x)$, the distribution of $L(x)$ asymptotically converges to the unit Gaussian distribution $p(L(x)) = \mathcal{N}[0,1]$ in the limit of large N according to the central limit theorem. Moreover, if the coefficients of $L(x)$ are sufficiently isotropic such that inequality (7) holds for $x$ sampled from $P_\Omega(x)$, $L(x)$ asymptotically follows a Gaussian distribution for large N.*

***Lemma 2 (Minimum cost among M samples).** Suppose that when $x$ is sampled from $P_\Omega(x)$, $L(x)$ follows a Gaussian distribution $p(L(x)) = \mathcal{N}[\mu, \sigma^2]$ with mean $\mu$ and variance $\sigma^2$. When the number of samples $M$ is sufficiently large, under Gaussian approximation, the minimum cost* *among these samples $L_M$ follows*

$$p_M(L_M) = \mathcal{N}\left[\mu - \sigma\sqrt{2\log M}, \sigma^2 \sqrt{\frac{\pi}{\log M}}\right] \quad (2)$$

*as the leading order.*

Although the derivation details are provided in the Methods section, these lemmas are briefly elucidated as follows: For Lemma 1, when $x$ follows $P_X(x)$, the weighted sum of the elements of $x$ as well as the products of $x$'s elements $x_{i_1} \cdots x_{i_\alpha}$ converges asymptotically to a Gaussian distribution according to the central limit theorem (**Fig. 1b**, bottom panel). For Lemma 2, we consider the probability that the minimum cost among $M$ samples does not fall below a particular $L_M$, which gives the cumulative distribution of $L_M$, $\varphi_M(L_M)$. Then, $p_M(L_M)$ is given as the derivative of $\varphi_M(L_M)$. The Gaussian approximation is employed to estimate the variance of $L_M$, which becomes accurate as $N$ increases owing to a sharp peak of $p_M(L_M)$.

Following these asymptotic properties, we characterise an approximate global minimum of $L(x)$. Because $X$ consists of $2^N$ states and $L(x)$ asymptotically follows the unit Gaussian distribution when $x$ follows $P_X(x)$ (Lemma 1), an approximate global minimum of $L(x)$, defined as $L_{\text{GM}} := \min[L(x^{(1)}), \ldots, L(x^{(k2^N)})]$, is computed using Lemma 2 as the case with $M = k2^N$. Thus, $L_{\text{GM}} \approx -\sqrt{2N\log 2}$ asymptotically holds for large $N$, where the effect of an order-one coefficient $k$ is negligibly smaller than the leading order. Numerical simulations show that the global minimum $L_{\text{GM}}$ follows $-\sqrt{2N\log 2}$, as predicted by equation (2) (**Fig. 1c**).

## Computational efficiency of gradient descent algorithms

In this section, we characterise the computational efficiency of gradient descent algorithms by explicitly computing the expected number of local minima $N_{LM}$ of cost functions. We define $D_i x$ as an operation that inverts the sign of an element of $x$, where the *i*th element is $D_i x_i = -x_i$ and the others are $D_i x_j = x_j$ for $i \neq j$. We also define the identical map $D_0 x = x$ for convenience. In this work, the gradient descent algorithm is defined as a sequence of operations that update a state $x_t \in X$ to minimise $L(x_t)$ by inverting the sign of only one element of $x_t$ for each step, given as $x_{t+1} = \underset{D_i x_t, i \in \{0,1,\ldots,N\}}{\operatorname{argmin}} L(D_i x_t)$. We refer to this algorithm as the gradient descent algorithm to emphasise an analogy to optimisation in continuous state space. When $x_{t+1} = x_t$, the search is finished, and this $x_t$ provides the local minimum of a basin wherein the initial state $x_0$ is involved.

Notably, the entire state space can be separated into basins, and each basin has a local minimum (**Fig. 2a**). Thus, the gradient descent algorithm is guaranteed to reach the local minimum when it starts from any position in a basin, where the number of convergence points of $x_t$ is $N_{LM}$ in the entire space. Crucially, the pigeonhole principle ensures that the gradient descent algorithm with random initialisation identifies the global minimum after a sufficient number of iterations. If the basin including the global minimum is of an average size, this approach requires $kN_{LM}$ iterations of random state initialisations. As the order-one coefficient *k* increases, the probability of a randomly selected initial state $x_0$ being placed in the basin of the global minimum at least once during the iterations converges to 1.

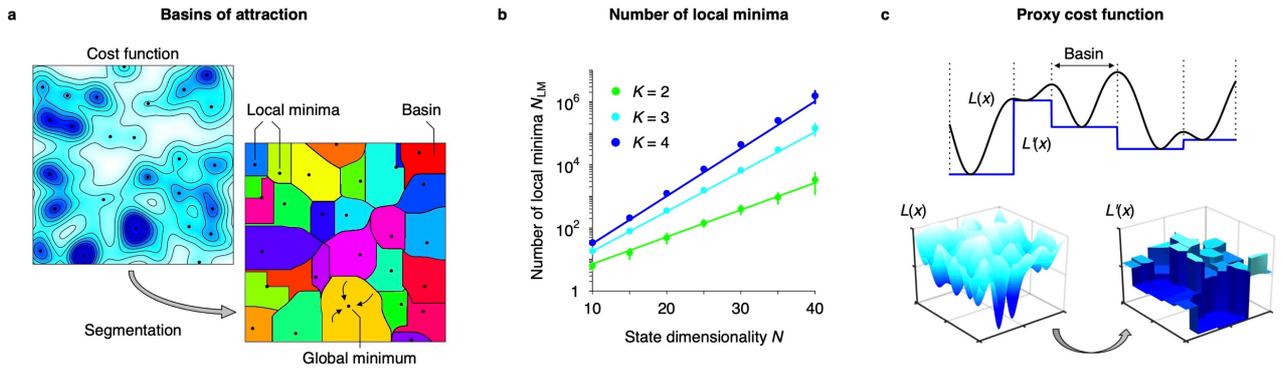

**Fig. 2. Computational efficiency of gradient descent algorithms. a,** Schematic of basins and local minima in the *N*-dimensional state space. In the left panel, darker colour represents lower cost. The entire space $X$ can be separated into basins (right panel), all of which have a local minimum (black dots). Thus, the efficiency of gradient descent algorithms depends on the number of local minima $N_{LM}$. **b,** Relationship between the state dimensionality *N* and the number of local minima $N_{LM}$. Green, cyan, and blue dots indicate the simulation results with *K* = 2, 3, 4, obtained with 20 different realisations of cost functions, and the error bars indicate the standard deviation; however, some error bars are hidden by the dots. The lines are theoretical predictions computed using equation (3). **c,** Original cost function *L*(*x*) (black) and proxy cost function *L'*(*x*) defined by the values of local minimum for each basin (blue). The lower graphs are 3D images of the cost function in **a** and the corresponding proxy cost function. The gradient descent algorithm identifies a local minimum of each basin within merely order *N* steps. Thus, the original problem can be replaced with the minimisation of *L'*(*x*) as a tractable proxy with a negligibly small computational cost. Although the gradient descent algorithm is no longer an efficient strategy for *L'*(*x*) because the gradient of *L'*(*x*) is zero for almost all states, this treatment makes it easier to combine the gradient descent algorithm with a non-local search algorithm (see the main text for details).

This property enables us to estimate the required computational cost for the gradient descent algorithm. To this end, we analytically compute the probability of a state $x$ sampled from $P_X(x)$ being a local minimum, $\text{Prob}[L(x) < L(D_1 x) \wedge \cdots \wedge L(x) < L(D_N x)]$, because the expectation of $N_{\text{LM}}$ is given as $N_{\text{LM}} = 2^N \text{Prob}[L(x) < L(D_1 x) \wedge \cdots \wedge L(x) < L(D_N x)]$. Considering a multidimensional binomial distribution with a weak correlation between elements, we obtain the following lemma:

***Lemma 3 (number of local minima).*** *For the class of cost functions defined in equation (1), the expected number of local minima is given as*

$$N_{\text{LM}} \approx \left(2 - \left(1 + \frac{2(\widehat{K} - 1)}{\pi}\right)^{-\frac{1}{2}}\right)^N \quad (3)$$

*as the order, where $\widehat{K}$ ($1 \leq \widehat{K} \leq K$) denotes the effective degree of $L(x)$.*

The derivation details and formal definition of $\widehat{K}$ are provided in the Methods section. Numerical simulations corroborate that the increase in the number of local minima matches equation (3) (**Fig. 2b**). Conversely, the number of steps required to reach a local minimum after starting from a random initial state is at most of the order *N*, owing to the binary state space. These observations indicate that the computational efficiency of the gradient descent algorithm with random initialisations is characterised by $N_{\text{LM}}$ because $N_{\text{LM}} \gg N$.

In short, we showed that naive gradient descent algorithms incur an exponential order iteration of random initialisations to visit the majority of basins for identifying an approximate global minimum because $N_{LM}$ is of an exponential order of $N$. This is unrealistic to achieve within a limited timeframe for large $N$, indicating the inefficiency of naive gradient descent approaches. Therefore, another solution is required.

To address this issue, in what follows, we consider a combination of the gradient descent algorithm and a non-local search algorithm. To make the analysis tractable, we consider the following operation: Replace the original cost function $L(x)$ with its local minima and define it as a proxy cost function $L'(x)$ (**Fig. 2c**). In other words, we define $L'(x)$, whose cost corresponds to the local minima of each basin of $L(x)$, that is, $L'(x) = \min\limits_{x \in \text{Basin}} L(x)$ for each basin. One may associate the transformation from $L(x)$ to $L'(x)$ with a low-pass filter that reduces the magnitude of the higher-degree products in $L(x)$. By considering this, we can include the outcome of the gradient descent algorithm in the definition of cost functions. Hereafter, we consider the optimisation of $L'(x)$ as well as $L(x)$.

**Computational efficiency of the optimal global search algorithm**

In this section, we derive an optimal strategy to minimise the computational cost for the global search of $L(x)$. To this end, we define a generic non-local search algorithm as an algorithm that combines the states experienced in the past to generate the subsequent (i.e., offspring) states (**Fig. 3a**). Generally, non-local search algorithms determine the subsequent states based on previous experiences because every information about $L(x)$ comes only from empirical observations. Intuitively, the offspring generated by the combination of preferable states are likely to be better than those generated randomly, as heuristically considered in evolutionary computation. Here, we formally demonstrate that this is the case for the class of cost functions considered herein.

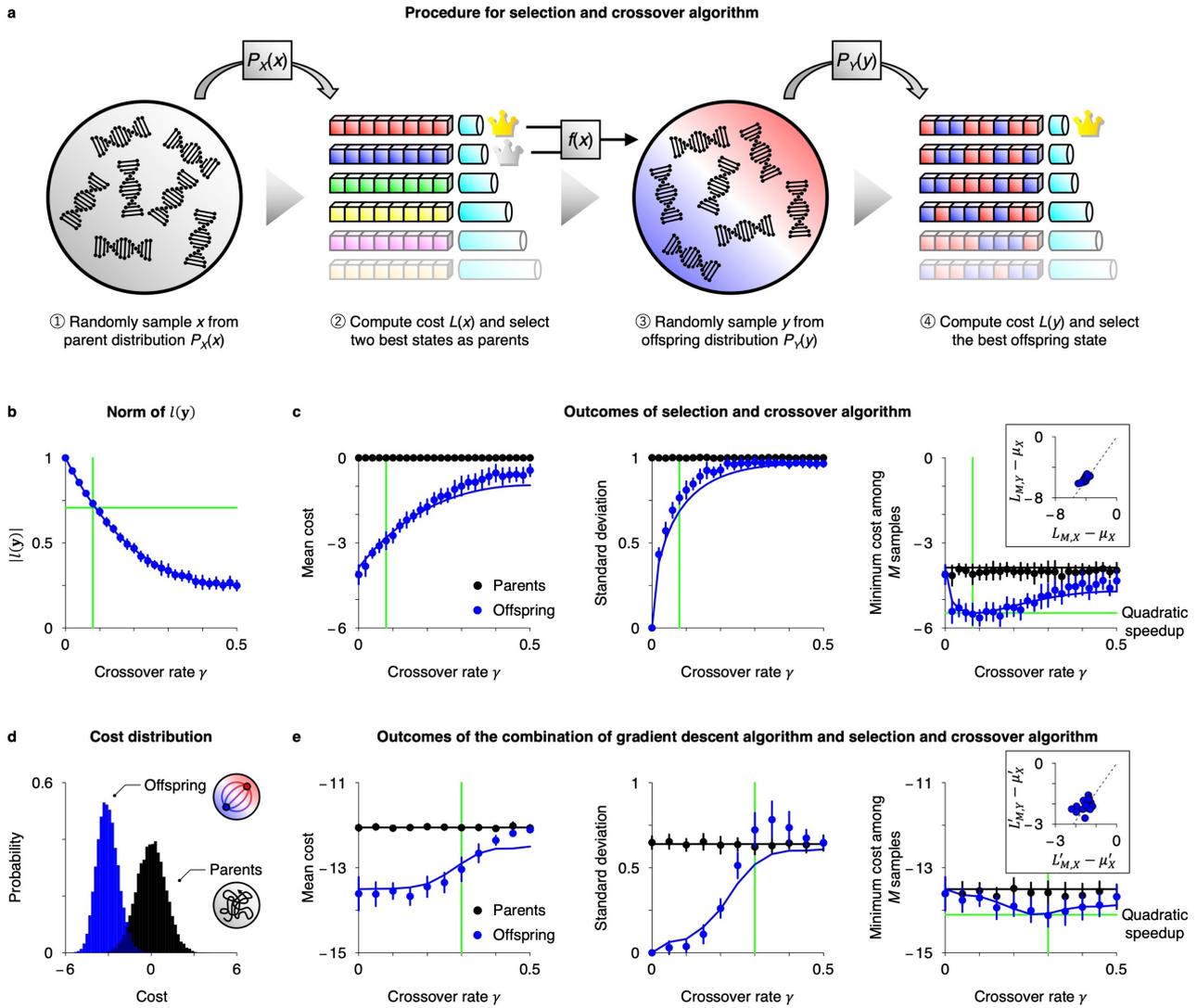

**Fig. 3. Schematics and outcomes of selection and crossover algorithm. a,** Schematic showing that a crossover of two favourable parent states generates offspring, where $f(x)$ determines a crossover weight distribution. **b,** Norm of a vector $l(\mathbf{y})$, which represents the components of $\mathbb{E}_{P_Y(y)}[L(y)]$. Dots are numerical results, whereas the line denotes the theoretical prediction obtained using equation (36) in the limit of large $N$. The green lines indicate the optimal norm

$|l(\mathbf{y})| = 1/\sqrt{2}$ and corresponding crossover rate $\gamma$ implicit in $l(\mathbf{y})$. **c,** Comparison of the parents' cost $L(x)$ (black) and offspring's cost $L(y)$ (blue). Panels show the mean, standard deviation, and minimum among *M* samples when the crossover rate is varied between $0 \leq \gamma \leq 0.5$. Fourth-degree cost functions in a 200-dimensional binary state space (*N* = 200, *K* = 4) are employed, and *M* = 20000 states are sampled for each condition. The blue lines are the theoretical predictions of the offspring's statistics, that is, mean $\mu_Y$, standard deviation $\sigma_Y$, and minimum among *M* samples $L_{M,Y}$ obtained using equation (6). Conversely, the black lines are the statistics of parents ($\mu_X$, $\sigma_X$, and $L_{M,X}$). The green lines indicate the optimal crossover rate $\gamma$ and the theoretically predicted optimal solution showing quadratic speedup (left panel). The inset panel compares the deviations of $L_{M,X}$ and $L_{M,Y}$ from $\mu_X$ when $\gamma$ is optimal, where the dashed line has a gradient of $\sqrt{2}$. **d,** Distributions of costs of parents (black) and offspring (blue) with the optimal $\gamma$. **e,** Same as **c**, but these panels show outcomes of the combination of the gradient descent algorithm and the selection and crossover algorithm. *M* = 100 states are sampled for each condition. For theoretical prediction, $|l(\mathbf{y})|$ of proxy cost function $L'(y)$ is estimated from numerically computed $\mu'_Y$ and $\sigma'_Y$, where the prime mark denotes the statistic of $L'(y)$. In **b, c,** and **e**, dots and error bars indicate the mean and standard deviation obtained by 20 different cost functions, respectively.

In this work, we define a non-local search algorithm as an algorithm that samples an offspring state $y = (y_1, ..., y_N)^T \in Y \subseteq X$ from a probability distribution $P_Y(y) = \prod_{i=1}^{N} P_{Y_i}(y_i)$ that is parameterised by a sequence of past states $x^{(1)}, ..., x^{(M)}$. The past states can be either randomly selected states or solutions of the gradient descent algorithm. Namely, $P_{Y_i}(y_i)$ is given as follows:

$$P_{Y_i}(y_i) = \begin{cases} \frac{1}{2} + \frac{1}{2} \sum_{m=1}^{M} f(x^{(m)}) x_i^{(m)} & \text{(for } y_i = 1) \\ \frac{1}{2} - \frac{1}{2} \sum_{m=1}^{M} f(x^{(m)}) x_i^{(m)} & \text{(for } y_i = -1) \end{cases} \quad (4)$$

where $\sum_{m=1}^{M} f(x^{(m)}) x_i^{(m)}$ is the weighted sum of the past states, and coefficient $f(x^{(m)})$ indicates a weight, which is cast as the probability of $x_i^{(m)}$ being selected as the offspring state $y_i$. Thus, the minimisation of the computational cost is reduced to the optimisation of $f(x)$.

Now, our aim is to identify the distribution $f(x)$ that minimises the minimum cost among $M$ states randomly sampled from the offspring distribution $P(y)$. The cost associated with $y$ is denoted as $L(y)$, which is defined by replacing $x$ in equation (1) with $y$. When the coefficients of $L(y)$ are sufficiently isotropic, given the Gaussianity of $L(y)$ (Lemma 1), the minimum cost among $M$ samples generated from $P(y)$ is estimated using Lemma 2. Thus, we define the objective function as follows:

$$\mathcal{L}(f(x)) \coloneqq \mathrm{E}_{P_Y(y)}[L(y)] - \sqrt{2 \log M \, \mathrm{Var}_{P_Y(y)}[L(y)]} \quad (5)$$

This comprises the mean $\mu_Y \coloneqq \mathrm{E}_{P_Y(y)}[L(y)]$ and variance $\sigma_Y^2 \coloneqq \mathrm{Var}_{P_Y(y)}[L(y)] = \mathrm{E}_{P_Y(y)}[L(y)^2] - \mathrm{E}_{P_Y(y)}[L(y)]^2$ of $L(y)$ over $P(y)$ following equation (2), where $\mathrm{E}_{P_Y(y)}[\cdot]$ denotes the expectation of $\cdot$ over $P(y)$. As equation (5) implies, here we focus on selection and crossover in a single generation for analytical tractability.

Our analysis shows that the mean and variance of $L(y)$ are approximated as $\mu_Y \approx |l(\mathbf{y})| \mathrm{E}_{f(x)}[L(x)]$ and $\sigma_Y^2 \approx 1 - |l(\mathbf{y})|^2$, respectively (see Methods for details). Here, $l(\mathbf{y}) \coloneqq$

$\mathrm{E}_{P_Y(y)}\left[(a_1 y_1, a_2 y_2, \ldots, a_{(N-K+1)\cdots N} y_{N-K+1} \cdots y_N)^{\mathrm{T}}\right]$ indicates a vector that arranges every component of $\mathrm{E}_{P_Y(y)}[L(y)]$ in each row, $|l(\mathbf{y})|$ denotes the norm of $l(\mathbf{y})$, and $\mathrm{E}_{f(x)}[L(x)]$ is the expectation of $L(x)$ over $f(x)$. Through a variational approach, we identify that $\mathcal{L}(f(x))$ is minimised when we assign the weights as $f(x^{(m_1)}) = 1 - \gamma$ for the best state that gives the minimum cost, $f(x^{(m_2)}) = \gamma$ for the second-best state, and $f(x^{(m)}) = 0$ otherwise, using a small positive value $\gamma$ representing the crossover rate ($0 < \gamma \leq 0.5$). This speaks to the selection and crossover algorithm with a biased crossover weight (**Fig. 3a**). According to Lemma 2, $L(x^{(m_1)}) \approx L(x^{(m_2)}) \approx -\sqrt{2 \log M}$ holds as the leading order. Therefore, we obtain the following theorem:

***Theorem 1 (quadratic speedup).*** *When the coefficients of $L(y)$ are sufficiently isotropic such that inequality (7) holds, from Lemmas 1 and 2, the minimum cost among M samples generated from the offspring distribution $P(y)$ is approximately lower bounded as follows:*

$$\mathcal{L}(f(x)) \approx -\sqrt{2 \log M}\left(|l(\mathbf{y})| + \sqrt{1 - |l(\mathbf{y})|^2}\right) \geq -\sqrt{2 \log M^2} \tag{6}$$

*The equality holds when and only when $|l(\mathbf{y})| = 1/\sqrt{2}$.*

The derivation details are provided in the Methods section. The theorem indicates that the optimal global search algorithm is a selection and crossover algorithm with a biased crossover rate $\gamma$ that satisfies $|l(\mathbf{y})| = 1/\sqrt{2}$. This is in significant contrast to usual selection and crossover algorithms in which the crossover weight is half and half. With this optimal $\gamma$, $\mathcal{L}(f(x)) \approx -\sqrt{2 \log M^2}$ is attained. Selection and crossover in a single generation is sufficient to exhibit this property. Remarkably, this scheme has the same efficiency as conventional gradient descent algorithms with $M^2$ random initialisations, meaning that—compared with conventional gradient descent algorithms—the optimal combination of the two algorithms quadratically speeds up the search efficiency better.

This indicates a substantial difference in their computational costs. As an idealised example setup, let us consider the minimisation of a cost function in a 1000-dimensional binary state space with an effective degree of $\hat{K} = 1.14$. When a computer conducts a trial of the gradient descent search per 0.1 ms, naive gradient descent algorithms incur approximately 2 million years to identify an approximate global minimum. Conversely, the optimal combination presented here can, in principle, reduce the required time to within one day, as long as the coefficients of the cost function are sufficiently isotropic.

We empirically corroborated our proposition using numerical simulations (**Fig. 3b–e**), which show that the present theorem (equation (6)) can predict the mean, standard deviation, and minimum of the offspring costs. First, we confirmed that the theoretical value of $|l(\mathbf{y})|$ obtained using equation (36) in the limit of large *N* precisely matches the numerical results (**Fig. 3b**). The costs of offspring are distributed with a mean substantially lower than the mean cost of the parent distribution (**Fig. 3c**, left). In contrast, the standard deviation of the offspring costs is close to that of the parents' costs, unless $\gamma$ is notably small (**Fig. 3c**, centre). Consequently, sampling from the offspring distribution leads to a higher chance of determining a better solution than sampling from the parent distribution (**Fig. 3c**, right). With the optimal $\gamma$, we observed that the minimum cost among *M* offspring states is $\sqrt{2}$ times negatively larger than the minimum cost among *M* parent states, indicating a quadratic speedup (**Fig. 3c**, inset). Conversely, the performance of the unbiased selection and crossover algorithm with $\gamma = 0.5$, which is widely used in the literature, was much poorer than that with the optimal $\gamma$. **Fig. 3d** visualises a substantial difference between the distributions of the parents' and offspring's costs when $\gamma$ is optimal.

In addition to a simple selection and crossover, we also confirmed that the combination of the gradient descent algorithm and the selection and crossover algorithm achieves the quadratic efficiency predicted by theory (**Fig. 3e**), where the theoretical prediction was conducted simply by replacing *L*(*x*) with *L'*(*x*). These results demonstrate the advantage of crossover between two favourable states over conventional gradient descent algorithms for the efficient global search.

In summary, we analytically and numerically demonstrated that a combination of the gradient descent algorithm and the selection and crossover algorithm can reduce the computational cost incurred for the global search to the square root order of that for conventional gradient descent algorithms. This owes to the properties that when the coefficients are sufficiently isotropic, the distribution of the offspring cost $L(y)$ is approximately cast as a Gaussian distribution (Lemma 1) and the minimum offspring cost is characterised by the sum of the mean and standard deviation of $L(y)$ over $P(y)$ (Lemma 2). Therefore, compared with conventional gradient descent approaches, the proposed combination with the optimal crossover rate can quadratically speed up the global search.

**Application to combinatorial optimisation problems**

Finally, we demonstrate the applicability of the quadratic speedup proposition to practical problems using the travelling salesman problem (TSP) as an example (**Fig. 4a**). TSP is a well-known combinatorial optimisation problem that incurs a huge computational cost to solve [24], wherein given a set of cities, the shortest route that enables all cities to be visited exactly once and then return to the starting city needs to be determined. Despite its straightforward definition, the problem becomes increasingly difficult when the number of cities increases.

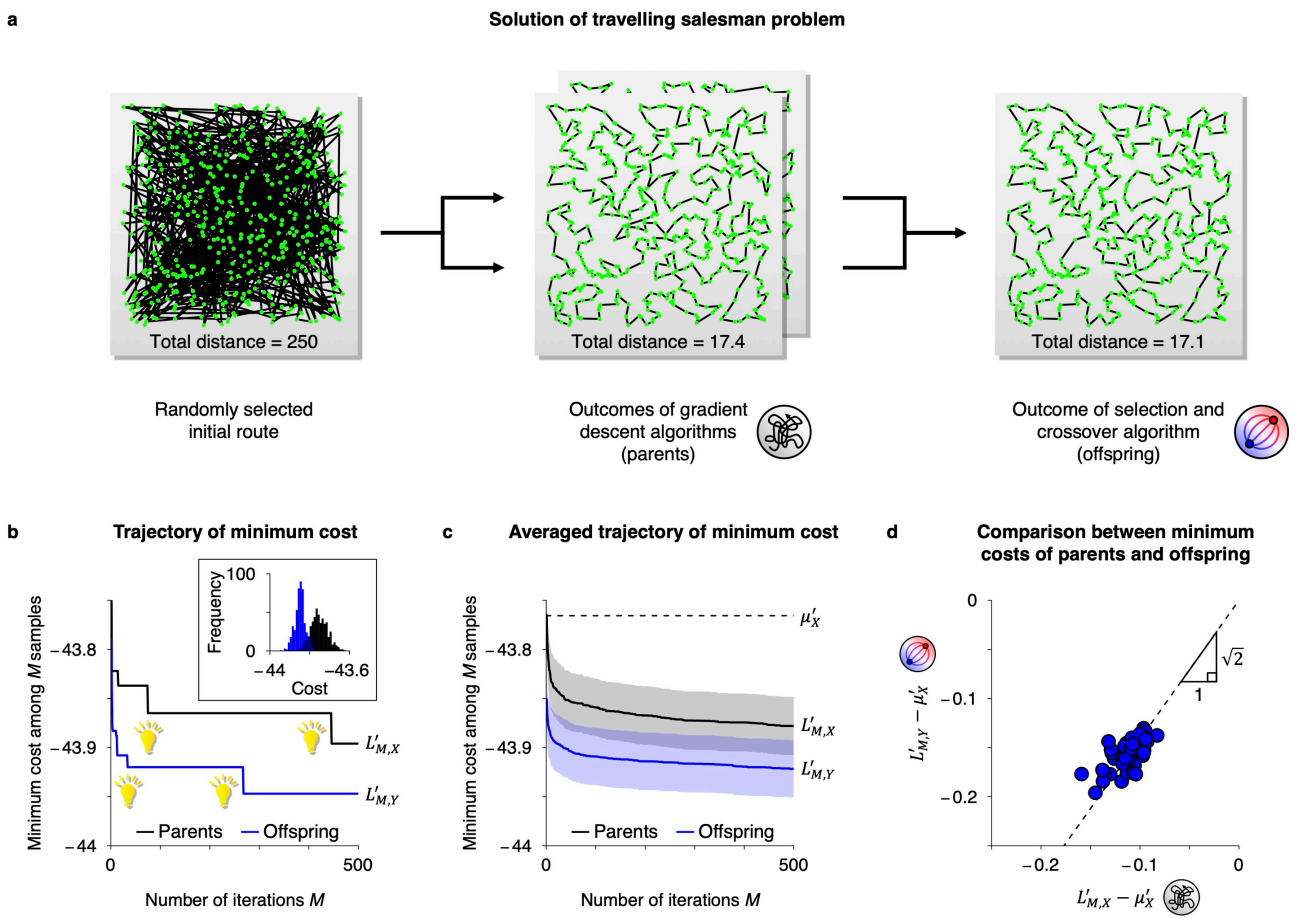

**Fig. 4. Application to the travelling salesman problem. a,** Schematic of the travelling salesman problem. The number of cities is $N$ = 500. The panels show randomly selected initial route (left), outcomes of the gradient descent algorithms (centre), and outcome of subsequent selection and crossover algorithm (right). **b,** Trajectory of minimum cost among $M$ samples when $M$ increases.

Here, optimisation of proxy cost function $L'(x)$ is considered. The black line indicates the parents' trajectory $L'_{M,X}$, whereas the blue line is the offspring trajectory $L'_{M,Y}$, which is substantially lower than the parents' one. The prime mark denotes the statistic of $L'(y)$. The crossover rate of $\gamma = 0.05$ is used. The inset panel depicts the overall cost distributions of parents (black) and offspring (blue). The cost function (i.e., total distance) was normalised prior to the optimisation to ensure zero mean and unit variance. **c,** Averaged trajectories obtained by 50 different travelling salesman problems. The shaded areas indicate the standard error. The dashed line indicates the parents' mean $\mu'_X$. **d,** Comparison between minimum costs of parents and offspring. In this plot, the horizontal axis indicates the difference between the parents' mean $\mu'_X$ and parents' minimum $L'_{M,X}$, whereas the vertical axis indicates the difference between $\mu'_X$ and the offspring minimum $L'_{M,Y}$. The dashed line has a gradient of $\sqrt{2}$.

Here, we define the gradient descent algorithm as an operation that repeats the exchange of the edges between two paths in a route to minimise the total distance, which is also known as the 2-opt algorithm [25]. Starting from a randomly selected initial route (**Fig. 4a**, left), the gradient descent algorithm identifies efficient routes (referred to as parents; **Fig. 4a**, centre). Then, we show that the subsequent selection and crossover—followed by the gradient descent of offspring—further reduce the cost, thereby facilitating the global search to identify a shorter route (**Fig. 4a**, right). During this process, the algorithm keeps the best route identified so far on the one hand and searches for a new route randomly. This yields a non-continuous trajectory of the minimum cost (**Fig. 4b**). Suddenly, the algorithm finds a better solution, and the cost is reduced immediately, similar to an aha moment or a eureka experience [26–29]. This implies a potential relation to problem solving with the insight that the brain performs (see Discussion for details). The different search efficiency between parents and offspring results in substantial differences in their minimum cost trajectories (**Fig. 4b**) and overall cost distributions (**Fig. 4b**, inset).

We qualitatively compared the efficiencies of the naive gradient descent algorithm and the present combination of the two algorithms. **Fig. 4c** depicts the minimum cost of parents and offspring identified within a limited timeframe (i.e., $M$ samples). The averaged trajectory of parents over various travelling salesman problems followed $-\sqrt{2 \log M}$, as predicted by Lemma 2. Compared with this, the offspring generated by the combination of the two algorithms could more efficiently identify a shorter route. The virtues of the selection and crossover algorithm are that it provides a better initial condition compared with the parents owing to a large negative mean cost, and that it can further reduce the cost efficiently owing to a large variance.

We further confirmed the quadratic speedup by showing the ratio between the deviation of the parents' and offspring's minimum costs from the parents' mean cost (**Fig. 4d**). We found that the latter was $\sqrt{2}$ times negatively larger than the former, indicating the quadratic speedup of the search efficiency from $M$ to $M^2$. Therefore, the proposed combination can efficiently identify a better solution within a limited timeframe. These results highlight the applicability of the proposed search scheme to practical optimisation problems, such as combinatorial optimisation problems.

**DISCUSSION**

Identifying an approximate global solution in generic optimisation problems incurs a huge computational cost owing to the presence of many local solutions, unless the problem can be cast as convex optimisation [30,31]. In this work, we showed that a combination of the gradient descent algorithm and the selection and crossover algorithm quadratically enhances the global search efficiency for a class of optimisation problems. Although the benefits of selection and crossover are often explained in terms of the building-block hypothesis [32,33] and the global search efficiency has been rigorously analysed for particular setups, such as the ONEMAX and JUMP functions [10–12] and the all-pairs shortest path problem [13], the conditions under which the selection and crossover algorithm speeds up the global search for generic cost functions remain to be established. To the best of our knowledge, this work is the first to demonstrate that a biased crossover of two preferable states (as opposed to an unbiased crossover) is the optimal non-local search strategy to maximise the global search efficiency for the class of cost functions considered herein.

The proposed efficient search scheme is associated with the efficiency in developing good ideas. Optimisation and problem solving can generally be formulated in terms of the minimisation of cost functions. For instance, in the case of video games or table games such as Shogi or Igo, the strategy for the next move is determined based on cost functions [34–36]. Finding the shortest route or solving mazes or puzzles is also performed by cost function minimisation, as we demonstrated. Indeed, generic inference and decision-making can be cast as cost function minimisation [21,22,37,38]; thus, the presented scheme has various potential applications. Furthermore, the proposed scheme can be combined with algorithms facilitating the gradient descent search [6] and other efficient search methods [39] that exploit the wisdom of deep learning and spin systems.

The applicability of the proposed theorem to practical problems is an important issue. Sufficient isotropy of coefficients is a criterion that constrains the degree to which an element of $x$ can be optimised without considering the majority of the other elements, which characterises the type of cost functions considered. This means that efficiency is guaranteed when the problem setup is less synergetic. In contrast, for highly structured cost functions that consist of high-degree products or strongly correlated coefficients, the present scheme might not largely improve the search efficiency. This is because the selection and crossover algorithm exploits the non-local correlation of costs implicit in the cost function, and highly synergetic cost functions may disturb it. In short, this scheme is not a solution for any problem without constraints that involves such a difficult setup. This is rather a mathematical truism according to the no free lunch theorem [1–4] because unless we set some conditions, there is no search strategy that is more efficient than others for any cost function. Thus, introducing the condition of sufficient isotropy is natural to specify the class of cost functions and derive the ensuing optimal search scheme.

The proposed theorem is potentially important for elucidating optimisations in biological organisms over multiple timescales. For instance, if a state $x$ represents a gene, the present scheme can be cast as a model of evolution, as considered in evolutionary computation [7–15] and neuroevolution [16–19]. Elucidating how evolution enables biological organisms to adapt to their environment is crucial to understand the intelligence [40,41]. Although biological verification is a difficult issue, if one can set up an experiment to compare the adaptation efficiencies of sexual and non-sexual selection—or strong and weak sexual selection [42]—it would be possible to examine if their efficiencies are quadratically different, as predicted by the proposed theorem. Thus, the present theory is, in principle, experimentally testable.

Moreover, brain activity—including neural activity and plasticity—can be explained as a process of minimising some cost function [21,22,37,38]. Thus, if a state $x$ represents neural activity and synaptic strengths, the present scheme can be cast as a computational model of the brain. In particular, it can be associated with an insight that enables to produce good ideas and solve problems efficiently [26–29,43]. Insight occurs when a solution is computed unconsciously and later emerges into awareness suddenly [28]. A previous work using pigeons suggested that an insightful search can solve task problems faster than a trial-and-error strategy [44]. In contrast to conventional learning algorithms in neurobiology that are generally modelled as local searches [45–49], the selection and crossover algorithm may unveil efficient optimisations beyond local searches that the brain may perform. One possible implementation in the neural network is that the brain may employ multiple generative models that represent external milieu states in different contexts or environments [50,51]. A new generative model can be created by combining two good generative models, presumably by changing synaptic connection structures [52,53]. This can quadratically speed up the search efficiency compared to naive synaptic plasticity. The validity of this hypothesis can potentially be examined by comparing with brain activity during insightful problem solving [26–29].

Furthermore, the proposed scheme can be combined with neuromorphic hardware [54,55], which will offer an insight-like computation that dramatically reduces the computational cost for global search in various tasks, while retaining its simple computational architecture. This is potentially important for a next-generation neuromorphic implementation to save energy and material cost as well as computational time.

Finally, it is noteworthy that the quadratic speedup presented herein has the same efficiency level as the Grover algorithm for quantum computers [56,57]. Surprisingly, even with a classical system, the present scheme can achieve an efficiency level offered by quantum algorithms for a class of optimisation problems.

In summary, we demonstrated that a combination of the gradient descent algorithm and the selection and crossover algorithm—with an optimal crossover bias—can reduce the computational cost to the square root order in contrast to naive gradient descent algorithms for a class of cost functions, facilitating the search efficiency quadratically. The virtue of the biased crossover of two favourable states is that it can provide offspring with a large negative cost and large diversity, where the crossover bias plays a key role in maximising efficiency. Owing to its simple computational architecture and minimal computational cost, the present scheme is highly desirable for optimisations in biological organisms over multiple timescales. It is also suitable for optimisations in biologically inspired computers such as neuromorphic hardware, offering the quadratically efficient global search strategy for artificial intelligence in various applications.

**METHODS**

**Sufficient isotropy of coefficients**

Let us suppose that $\omega = (\omega_1, \ldots, \omega_N)^T \in \Omega \subseteq X$ is a random binary state sampled from a probability distribution $P_\Omega(\omega) = \prod_{i=1}^N P_{\Omega_i}(\omega_i)$. We denote the expectation of $\omega$ as $\boldsymbol{\omega} = \mathrm{E}_{P_\Omega(\omega)}[\omega] = \int \omega P_\Omega(\omega) d\omega$ using boldface. Moreover, we define a vector expression of $L(\omega)$'s components as follows: $l(\omega) := (a_1\omega_1, a_2\omega_2, \ldots, a_{(N-K+1)\cdots N}\omega_{N-K+1}\cdots \omega_N)^\mathrm{T}$, which indicates a vector that arranges all components of $L(\omega)$ in each row, where $L(\omega)$ is defined by replacing $x$ in equation (1) with $\omega$. Here, the number of all components of $L(\omega)$, denoted as $n$, is given as the sum of the binomial coefficients as follows: $n = \sum_{\alpha=1}^K \binom{N}{\alpha}$, which is of order $N^K$. By construction, $L(\omega) = \sum_{i=1}^n l(\omega)_i$ is satisfied, where $l(\omega)_i$ with $i \in \{1,2,\ldots,n\}$ indicates the $i$th component of $l(\omega)$. A vector expression for the expectation of $L(\omega)$, $\mathrm{E}_{P_\Omega(\omega)}[L(\omega)]$, is given as $l(\boldsymbol{\omega}) = \mathrm{E}_{P_\Omega(\omega)}[l(\omega)]$. Moreover, the norm of $l(\omega)$ is defined as $|l(\omega)| := \sqrt{\sum_{i=1}^n l(\omega)_i^2} = \sqrt{\sum_{i=1}^N a_i^2 + \sum_{i<j} a_{ij}^2 + \cdots}$. The norm of $l(x)$ as well as that of $l(\omega)$ is 1 because we define the variance of $L(x)$ as 1 when $x$ is sampled from $P_X(x)$ (see the Results section). Conversely, the norm of $l(\boldsymbol{\omega})$ is between 0 and 1. The deviation of $l(\omega)$ from $l(\boldsymbol{\omega})$ is denoted as $\Delta l(\omega) := l(\omega) - l(\boldsymbol{\omega})$.

In this work, we consider that the coefficients of $L(\omega)$, $\{a_i, a_{ij}, \ldots, a_{i_1\cdots i_K}\}$, are of the same order. Moreover, we assume that the coefficients of $L(\omega)$ are sufficiently isotropic when they satisfy the following inequalities for almost all random variables $\omega$ sampled from $P_\Omega(\omega)$:

$$\begin{cases} \left| \sum_{i \in C_1(N^{K-1})} \Delta l(\omega)_i \right| \leq \mathcal{O}(1) \\ \left| \sum_{(i,i') \in C_2(N^{2K-1})} \Delta l(\omega)_i \Delta l(\omega)_{i'} \right| < \mathcal{O}(N^{-\varepsilon}) \end{cases} \quad (7)$$

where $\mathcal{O}(N^{-\varepsilon})$ indicates an order $N^{-\varepsilon}$ value with a sufficiently small positive constant $\varepsilon$. Here, $C_1(N^{K-1}) \subseteq \{1,2,\ldots,n\}$ indicates an arbitrary subset of indices $i$, where the number of elements in $C_1(N^{K-1})$ is of order $N^{K-1}$. Moreover, $C_2(N^{2K-1})$ indicates an arbitrary subset of pairs of $(i, i')$ that satisfy $i \neq i'$ and $1 \leq i, i' \leq n$, where the number of elements in $C_2(N^{2K-1})$ is of order $N^{2K-1}$. This condition serves to ensure that the covariance of a small subset of $l(\omega)$'s elements is sufficiently smaller than the variance of $L(\omega)$ when N is large, which implies that the coefficients are isotropically distributed over elements of $l(\omega)$.

**Derivation of Lemma 1**

Lemma 1 is a corollary of the central limit theorem. First, we consider the Gaussianity of $L(x)$ when $x$ is sampled from $P_X(x)$. The characteristic function of $L(x)$ and its logarithm are defined as $\varphi(\tau) := \mathrm{E}_{P_X(x)}[e^{i\tau L(x)}]$ and $\Phi(\tau) := \log \varphi(\tau)$, respectively. The first- to fourth-order derivatives of $\Phi(\tau)$ with respect to $\tau$ are computed as follows:

$$\begin{cases} \Phi^{(1)}(\tau) = \mathrm{E}_{P_X(x)}[iL(x)e^{i\tau L(x)}]/\varphi(\tau) \\ \Phi^{(2)}(\tau) = \mathrm{E}_{P_X(x)}[-L(x)^2 e^{i\tau L(x)}]/\varphi(\tau) - \Phi^{(1)}(\tau)^2 \\ \Phi^{(3)}(\tau) = \mathrm{E}_{P_X(x)}[-iL(x)^3 e^{i\tau L(x)}]/\varphi(\tau) - 3\Phi^{(1)}(\tau)\Phi^{(2)}(\tau) - \Phi^{(1)}(\tau)^3 \\ \Phi^{(4)}(\tau) = \mathrm{E}_{P_X(x)}[L(x)^4 e^{i\tau L(x)}]/\varphi(\tau) - \Phi^{(1)}(\tau)\mathrm{E}_{P_X(x)}[-iL(x)^3 e^{i\tau L(x)}]/\varphi(\tau) \\ \qquad\qquad -3\Phi^{(2)}(\tau)^2 - 3\Phi^{(1)}(\tau)\Phi^{(3)}(\tau) - 3\Phi^{(1)}(\tau)^2\Phi^{(2)}(\tau) \end{cases} \qquad (8)$$

For $\tau = 0$, we obtain

$$\begin{cases} \Phi(0) = \log \mathrm{E}_{P_X(x)}[1] = 0 \\ \Phi^{(1)}(0) = i\mathrm{E}_{P_X(x)}[L(x)] = 0 \\ \Phi^{(2)}(0) = -\mathrm{E}_{P_X(x)}[L(x)^2] = -1 \\ \Phi^{(3)}(0) = -i\mathrm{E}_{P_X(x)}[L(x)^3] \\ \Phi^{(4)}(0) = \mathrm{E}_{P_X(x)}[L(x)^4] - 3 \end{cases} \qquad (9)$$

Here, $\mathrm{E}_{P_X(x)}[L(x)] = 0$ and $\mathrm{E}_{P_X(x)}[L(x)^2] = 1$ hold because we defined them as such (see the Results section). Owing to the symmetry of $L(x)$ with respect to $P_X(x)$, $\mathrm{E}_{P_X(x)}[L(x)^3] = 0$ holds. Because each element of $x$ is independently sampled from $P_X(x)$ and its expectation is zero, $\mathrm{E}_{P_X(x)}[L(x)^4]$ is computed as follows:

$$\mathrm{E}_{P_X(x)}[L(x)^4] = \mathrm{E}_{P_X(x)}\left[\sum_{i=1}^{n} l(x)_i^4 + 3\sum_{i=1}^{n}\sum_{i\neq i'} l(x)_i^2 l(x)_{i'}^2\right] = 3 - 2\mathrm{E}_{P_X(x)}\left[\sum_{i=1}^{n} l(x)_i^4\right] \quad (10)$$

Here, $\sum_{i=1}^{n} l(x)_i^2 = 1$ was used to obtain the last equality. When the coefficients of $L(x)$ are of the same order, $\mathrm{E}_{P_X(x)}[\sum_{i=1}^{n} l(x)_i^4]$ is of order $n^{-1}$, which converges to zero when *N* increases. Thus, the fourth-order cumulant $\Phi^{(4)}(0)$ asymptotically converges to zero for large *N*. This also holds true for the higher order cumulants $\Phi^{(k)}(0)$ with $k \geq 5$. Hence, the Taylor expansion of $\Phi(\tau)$ can be expressed as

$$\Phi(\tau) = \sum_{k=0}^{\infty} \frac{\Phi^{(k)}(0)}{k!}\tau^k = -\frac{\sigma_X^2}{2}\tau^2 + \mathcal{O}(n^{-1}) \quad (11)$$

using $\sigma_X^2 = \mathrm{E}_{P_X(x)}[L(x)^2] = 1$. Therefore, from the inversion theorem, we obtain

$$p(L(x)) = \lim_{N\to\infty} \frac{1}{2\pi}\int_{-\infty}^{\infty} e^{-i\tau L(x)}\varphi(\tau)d\tau = \frac{1}{\sqrt{2\pi\sigma_X^2}} e^{-\frac{L(x)^2}{2\sigma_X^2}} \quad (12)$$

in the limit of large *N*. Thus, $p(L(x))$ asymptotically converges to the unit Gaussian distribution.

Next, we show the Gaussianity of $L(\omega)$ over $P_\Omega(\omega)$ when inequality (7) is satisfied. The mean of $L(\omega)$, $\mathrm{E}_{P_\Omega(\omega)}[L(\omega)]$, is given as the element sum of $l(\boldsymbol{\omega})$, denoted as $L(\boldsymbol{\omega}) = \sum_{i=1}^{n} l(\boldsymbol{\omega})_i = \mathrm{E}_{P_\Omega(\omega)}[L(\omega)]$, which is equal to the value of equation (1) when $\boldsymbol{\omega}$ is substituted. Moreover, the variance of $L(\omega)$, $\mathrm{Var}_{P_\Omega(\omega)}[L(\omega)] = \mathrm{E}_{P_\Omega(\omega)}\left[(L(\omega) - L(\boldsymbol{\omega}))^2\right] = \mathrm{E}_{P_\Omega(\omega)}[\sum_{i=1}^{n}\sum_{i'=1}^{n} \Delta l(\omega)_i \Delta l(\omega)_{i'}]$, is computed as follows: When the coefficients of $L(\omega)$ are sufficiently isotropic to satisfy inequality (7), the sum of the terms over unmatched indices $i \neq i'$ is negligibly small. Namely, it satisfies $\left|\mathrm{E}_{P_\Omega(\omega)}[\sum_{i=1}^n \sum_{i \neq i'} \Delta l(\omega)_i \Delta l(\omega)_{i'}]\right| < \mathcal{O}(N^{-\varepsilon})$ according to inequality (7). This holds true because for the $\alpha$th-degree products, $\binom{N-\alpha}{\alpha}$ terms have non-overlapped indices that satisfy $i_1 \neq i_1', \ldots, i_\alpha \neq i_\alpha'$, and thus their covariances are zero, that is, $\mathrm{E}_{P_\Omega(\omega)}[(\omega_{i_1} \cdots \omega_{i_\alpha} - \boldsymbol{\omega}_{i_1} \cdots \boldsymbol{\omega}_{i_\alpha})(\omega_{i_1'} \cdots \omega_{i_\alpha'} - \boldsymbol{\omega}_{i_1'} \cdots \boldsymbol{\omega}_{i_\alpha'})] = 0$. Thus, for a $K$th-degree cost functions, only $\mathcal{O}(N^{2K-1})$ terms have non-zero values. Their sum is upper bounded by the order $N^{-\varepsilon}$ owing to inequality (7). For instance, when the coefficients are independently sampled from a probability distribution, $\mathrm{E}_{P_\Omega(\omega)}[\sum_{i=1}^n \sum_{i \neq i'} \Delta l(\omega)_i \Delta l(\omega)_{i'}]$ is of order $N^{-1/2}$, which is much smaller than order $N^{-\varepsilon}$ for large $N$. Hence, when the coefficients of $L(\omega)$ are sufficiently isotropic, we obtain

$$\mathrm{Var}_{P_\Omega(\omega)}[L(\omega)] = \mathrm{E}_{P_\Omega(\omega)}\left[\sum_{i=1}^n \Delta l(\omega)_i^2\right] + \mathcal{O}(N^{-\varepsilon})$$

$$= \sum_{i=1}^N a_i^2(1 - \boldsymbol{\omega}_i^2) + \sum_{i<j} a_{ij}^2(1 - \boldsymbol{\omega}_i^2 \boldsymbol{\omega}_j^2) + \cdots + \sum_{i_1 < \cdots < i_K} a_{i_1 \cdots i_K}^2 (1 - \boldsymbol{\omega}_{i_1}^2 \cdots \boldsymbol{\omega}_{i_K}^2) + \mathcal{O}(N^{-\varepsilon})$$

$$= 1 - |l(\boldsymbol{\omega})|^2 + \mathcal{O}(N^{-\varepsilon}) \tag{13}$$

Thus, $\mathrm{Var}_{P_\Omega(\omega)}[L(\omega)] = 1 - |l(\boldsymbol{\omega})|^2$ is obtained in the limit of large $N$.

Furthermore, the higher order cumulants can be computed by replacing $L(x)$ in equation (9) with the deviation $\Delta L(\omega) \coloneqq L(\omega) - L(\boldsymbol{\omega}) = \sum_{i=1}^n \Delta l(\omega)_i$, which enables us to assess the characteristic function of $L(\omega)$. For the third-order cumulant, we obtain

$$\mathrm{E}_{P_\Omega(\omega)}[\Delta L(\omega)^3] = \mathrm{E}_{P_\Omega(\omega)}\left[\sum_{i=1}^n \Delta l(\omega)_i^3 + \sum_{(i,i',i'') \in \mathrm{PO}} \Delta l(\omega)_i \Delta l(\omega)_{i'} \Delta l(\omega)_{i''}\right] \tag{14}$$

The first term is the sum over identical indices, which is negligibly smaller than 1 when the coefficients of $L(\omega)$ are of the same order. The second term is the sum over triplets of indices with which elements of $\omega$ are partially overlapped (PO) with each other, which is assessed as

$$\sum_{(i,i',i'')\in \text{PO}} \Delta l(\omega)_i \Delta l(\omega)_{i'} \Delta l(\omega)_{i''} = \sum_{i \in C_1(N^{K-1})} \Delta l(\omega)_i \cdot \sum_{(i',i'') \in C_2(N^{2K-1})} \Delta l(\omega)_{i'} \Delta l(\omega)_{i''} < \mathcal{O}(N^{-\varepsilon})$$

according to inequality (7). Conversely, the expectation of the sum over the index triplets with which elements of $\omega$ are not overlapped is zero. Therefore, we find that $\left|\mathrm{E}_{P_\Omega(\omega)}[\Delta L(\omega)^3]\right| < \mathcal{O}(N^{-\varepsilon})$, which asymptotically converges to zero for large $N$.

The same strategy can be applied to assess the higher order cumulants $\Phi^{(k)}(0)$ with $k \geq 4$. Hence, similar to equation (11), we obtain

$$\Phi(\tau) = -\frac{\sigma_\Omega^2}{2}\tau^2 + \mathcal{O}(N^{-\varepsilon}) \tag{15}$$

with $\sigma_\Omega^2 = 1 - |l(\boldsymbol{\omega})|^2$. Therefore, by applying equation (12), we obtain that $L(\omega)$ asymptotically follows the following Gaussian distribution

$$p(L(\omega)) = \mathcal{N}[L(\boldsymbol{\omega}), 1 - |l(\boldsymbol{\omega})|^2] \tag{16}$$

in the limit of large $N$.

**Derivation of Lemma 2**

The integral of $p(L)$ provides the cumulative distribution function of $L \equiv L(x)$, $\varphi(L) \coloneqq \int_{-\infty}^{L} p(L')dL'$, where $0 \leq \varphi(L) \leq 1$. Moreover, the cumulative distribution function of the minimum cost among $M$ samples $L_M$ is given as $\varphi_M(L_M) \coloneqq 1 - \text{Prob}[L_M < L(x^{(1)}) \wedge \cdots \wedge L_M < L(x^{(M)})]$. Because $x^{(1)}, \ldots, x^{(M)}$ are sampled in a mutually independent manner, $\varphi_M(L_M) = 1 - $ $\prod_{m=1}^{M} \text{Prob}[L_M < L(x^{(m)})] = 1 - (1 - \varphi(L_M))^M$ holds. By construction, the derivative of

$\varphi_M(L_M)$ yields the probability density of $L_M$, given as

$$p_M(L_M) = M(1 - \varphi(L_M))^{M-1} p(L_M) \qquad (17)$$

Because $p_M(L_M)$ is a unimodal distribution with a sharp peak at the mode, $p_M(L_M)$ can be approximated as a Gaussian distribution, where the mean of $p_M(L_M)$ matches its mode.

The mode of $p_M(L_M)$ is given by the following zero-gradient condition:

$$p'_M(L_M) = -M(M-1)(1 - \varphi(L_M))^{M-2} p(L_M)^2 + M(1 - \varphi(L_M))^{M-1} p'(L_M) = 0 \qquad (18)$$

where $p'(L_M) = \left.\frac{dp(L)}{dL}\right|_{L=L_M}$ is the derivative of $p(L)$ at $L = L_M$. Thus, the mode satisfies

$(M-1)p(L_M)^2 = (1 - \varphi(L_M))p'(L_M)$, which is approximated as $Mp(L_M) \simeq \frac{p'(L_M)}{p(L_M)}$ owing to

$M \gg 1$ and $\varphi(L_M) \ll 1$. Substituting $p(L) = \mathcal{N}[\mu, \sigma^2]$ into the last equality and taking the logarithm yield

$$\log M - \frac{(L_M - \mu)^2}{2\sigma^2} - \frac{1}{2}\log(2\pi\sigma^2) \simeq \log\left(-\frac{L_M - \mu}{\sigma^2}\right) \qquad (19)$$

Hence, by assuming that $\log\left(-\frac{L_M - \mu}{\sigma^2}\right) + \frac{1}{2}\log(2\pi\sigma^2)$ is sufficiently smaller than the other terms for large $M$, we obtain the mode of $p_M(L_M)$ as follows:

$$L_M = \mu - \sigma\sqrt{2\log M} \qquad (20)$$

as the leading order. We confirmed that $\log\left(-\frac{L_M - \mu}{\sigma^2}\right) + \frac{1}{2}\log(2\pi\sigma^2)$ is of order $\log(\log M)$, which is negligibly smaller than the other terms; thus, the above assumption holds true for $M \gg 1$. We note that by substituting equation (20) into the right-hand side of equation (19), the first-order perturbation of $L_M$ is obtained as follows:

$$L_M = \mu - \sigma\sqrt{2\log M - \log(4\pi \log M)} \tag{21}$$

which provides a more precise estimation of $L_M$ (see **Fig. 1c**).

Moreover, the variance of $p_M(L_M)$ is computed as follows: Employing the Gaussian approximation $p_M(L_M) = \mathcal{N}[\mu_M, \sigma_M^2]$, the second-order Taylor expansion of $-\log p_M(L_M)$ around mean $\mu_M$ yields

$$-\log p_M(L_M) = -\log p_M(\mu_M) + \frac{1}{2\sigma_M^2}(L_M - \mu_M)^2 + \mathcal{O}((L_M - \mu_M)^3) \tag{22}$$

The Hessian of $-\log p_M(L_M)$, also known as the Fisher information, gives $\sigma_M^{-2}$. Thus, when $M$ is large, from equations (17) and (18), we obtain

$$\frac{1}{\sigma_M^2} = -\frac{d^2 \log p_M(L_M)}{dL_M^2}\bigg|_{L_M=\mu_M} = -\frac{d}{dL_M}\frac{p'_M(L_M)}{p_M(L_M)}\bigg|_{L_M=\mu_M}$$

$$= -(M-1)\left(-(1-\varphi(\mu_M))^{-2}p(\mu_M)^2 + (1-\varphi(\mu_M))^{-1}p'(\mu_M)\right) - \frac{1}{\sigma^2}$$

$$\approx -\frac{M}{\sigma^2}(\mu_M - \mu)p(\mu_M) - \frac{1}{\sigma^2}$$

$$= \frac{M\sqrt{2\log M}}{\sigma}\frac{1}{\sqrt{2\pi\sigma^2}}\exp\left(-\frac{(\sigma\sqrt{2\log M})^2}{2\sigma^2}\right) - \frac{1}{\sigma^2}$$

$$\approx \frac{1}{\sigma^2}\sqrt{\frac{\log M}{\pi}} \tag{23}$$

Therefore, from $\sigma_M^2 = \sigma^2\sqrt{\pi/\log M}$, we obtain Lemma 2 for large $M$.

**Derivation of Lemma 3**

The expected number of local minima $N_{\text{LM}}$ is obtained as the product of the probability of a state being a local minimum, $\text{Prob}[L(x) < L(D_1 x) \wedge \cdots \wedge L(x) < L(D_N x)]$, and the number of all possible states, $2^N$. First, we define the sum of all $x_i$-related terms in $L(x)$ as

$$b_i := a_i x_i + \sum_{j \neq i} a_{ij} x_i x_j + \cdots + \sum_{\substack{i \neq i_2, \ldots, i \neq i_K \\ 1 \leq i_2 < \cdots < i_K \leq N}} a_{i i_2 \cdots i_K} x_i x_{i_2} \cdots x_{i_K} \quad (24)$$

Because $D_i x_i = -x_i$ and $D_i x_j = x_j$ for $i \neq j$, we have $x_i - D_i x_i = 2 x_i$ and $x_j - D_i x_j = 0$. Thus, the condition $L(x) < L(D_i x)$ is reduced to $b_i < 0$.

Here, $b_i$ can be cast as the considered class of cost functions. This means that, when $x$ is sampled from $P_X(x)$, $b_i$ asymptotically follows a zero-mean Gaussian distribution $p_b(b_i) = \mathcal{N}[0, \sigma_{b_i}^2]$ for large $N$ according to Lemma 1, where $\sigma_{b_i}^2 = \text{Var}_{P_X(x)}[b_i] = \text{E}_{P_X(x)}[b_i^2] - \text{E}_{P_X(x)}[b_i]^2$ denotes the variance of $b_i$. Here, we assume that $a_i x_i$ is negligibly small relative to the sum of the other terms in $b_i$ when $N$ is large.

The condition $b_i < 0$ is further rewritten using the Heaviside step function $\Theta(b_i)$. Because $\Theta(b_i) = 0$ for $b_i < 0$ and $\Theta(b_i) = 1$ otherwise, by defining a new effective variable $c := \sum_{i=1}^{N} \Theta(b_i)$, the probability of $x$ being a local minimum is cast as the probability of $c$ being zero; that is, $\text{Prob}[L(x) < L(D_1 x) \wedge \cdots \wedge L(x) < L(D_N x)] = P(c = 0)$. Because $c$ indicates the number of non-negative components in $b_1, \ldots, b_N$, its probability is characterised in the form of a binomial distribution as follows:

$$P(c) = \frac{1}{Z} \binom{N}{c} q(c)^c (1 - q(c))^{N-c} \quad (25)$$

where $q(c)$ denotes the probability of $b_i$ being positive for a given $c$, and $Z$ denotes the partition function.

In contrast to the usual binomial distributions, here $q(c)$ is a function of $c$ because $b_1, \ldots, b_N$ are weakly correlated with each other. The following properties of $q(c)$ are observed: (1) when the number of positive components equals that of negative components, the correlation is negligible, meaning that $q(c = N/2) = 1/2$; (2) owing to the symmetry of $p(L(x))$, $P(c)$ is a symmetric distribution with respect to $c = N/2$, which leads to the symmetry of $q(c)$ with respect to $c = N/2$, $q(c) = 1 - q(N - c)$; and (3) $q(c)$ is a monotonically increasing function of $c$, and its gradient is sufficiently small relative to N because $0 \leq q(c) \leq 1$. Considering these properties, we adopt the following linear approximation of $q(c)$:

$$q(c) = \frac{1}{2} + \lambda \left( \frac{c}{N} - \frac{1}{2} \right) \tag{26}$$

where $\lambda$ ($0 \leq \lambda \leq 1$) indicates the correlation strength that is determined by the magnitude of the correlations among $b_1, \ldots, b_N$. In particular, when $c = 0$, substituting equation (26) into equation (25) gives the probability of $x$ being a local minimum up to unknown parameters $\lambda$ and $Z$ as follows:

$$P(c = 0) = \frac{1}{Z} \left( \frac{1 + \lambda}{2} \right)^N \tag{27}$$

To further characterise parameters $\lambda$ and $Z$, we next approximate $P(c)$ using Stirling's formula, $n! \approx \sqrt{2\pi n} \left( \frac{n}{e} \right)^n$. The binomial coefficient is approximated as

$$\binom{N}{c} \approx \sqrt{\frac{N}{2\pi c(N - c)}} \frac{1}{\left( \frac{c}{N} \right)^c \left( 1 - \frac{c}{N} \right)^{N-c}} \tag{28}$$

Thus, substituting equations (26) and (28) into equation (25) and using $n^{n'} = e^{n' \log n}$, we obtain

$$P(c) = \frac{1}{Z}\sqrt{\frac{N}{2\pi c(N-c)}} \frac{\left(1+\lambda\left(\frac{2c}{N}-1\right)\right)^c \left(1-\lambda\left(\frac{2c}{N}-1\right)\right)^{N-c}}{\left(1+\left(\frac{2c}{N}-1\right)\right)^c \left(1-\left(\frac{2c}{N}-1\right)\right)^{N-c}}$$

$$= \frac{1}{Z}\sqrt{\frac{N}{2\pi c(N-c)}} \exp\left[c\left\{\log\left(1+\lambda\left(\frac{2c}{N}-1\right)\right) - \log\left(1+\left(\frac{2c}{N}-1\right)\right)\right\}\right.$$

$$\left. +(N-c)\left\{\log\left(1-\lambda\left(\frac{2c}{N}-1\right)\right) - \log\left(1-\left(\frac{2c}{N}-1\right)\right)\right\}\right] \quad (29)$$

According to the central limit theorem, when $N$ is large and $|c - N/2| \ll N/2$, $P(c)$ asymptotically converges to a Gaussian distribution. Using the Taylor expansion $\log\left(1+\left(\frac{2c}{N}-1\right)\right) = \left(\frac{2c}{N}-1\right) - \frac{1}{2}\left(\frac{2c}{N}-1\right)^2 + \mathcal{O}\left(\left(\frac{2c}{N}-1\right)^3\right)$, we obtain

$$P(c) = \frac{1}{Z}\sqrt{\frac{2}{\pi N}} e^{-\frac{2(1-\lambda)^2}{N}\left(c-\frac{N}{2}\right)^2 + \mathcal{O}\left(\left(c-\frac{N}{2}\right)^3\right)} \quad (30)$$

for $|c - N/2| \ll N/2$. This indicates that $P(c)$ asymptotically converges to a Gaussian distribution, $\mathcal{N}\left[\frac{N}{2}, \frac{N}{4(\lambda-1)^2}\right] = \sqrt{\frac{2(1-\lambda)^2}{\pi N}} e^{-\frac{2(1-\lambda)^2}{N}\left(c-\frac{N}{2}\right)^2}$. Owing to this asymptotic relationship, the partition function is given as $Z = (1-\lambda)^{-1}$, and the variance of $c$ is given as $\mathrm{Var}_{P_X(x)}[c] = \frac{N}{4}(1-\lambda)^{-2}$.

This means that $\lambda$ and $Z$ can be identified by computing $\mathrm{Var}_{P_X(x)}[c]$ because $\lambda = 1 - \left(4\mathrm{Var}_{P_X(x)}[c]/N\right)^{-1/2}$. Thus, we now estimate $\mathrm{Var}_{P_X(x)}[c]$ as follows: When $b_i$ and $b_j$ are weakly correlated Gaussian variables, their arbitrary nonlinear functions, $g(b_i)$ and $g(b_j)$, are also weakly correlated. Based on the Taylor expansion with respect to the covariance $\mathrm{Cov}_{P_X(x)}[b_i, b_j] = \mathrm{E}_{P_X(x)}[b_i b_j] - \mathrm{E}_{P_X(x)}[b_i]\mathrm{E}_{P_X(x)}[b_j]$, previous work [23,58] showed that

$$\mathrm{Cov}_{P_X(x)}[g(b_i), g(b_j)] = \sum_{n=1}^{\infty} \frac{\mathrm{E}_{P_X(x)}[g^{(n)}(b_i)] \mathrm{E}_{P_X(x)}[g^{(n)}(b_j)]}{n!} \mathrm{Cov}_{P_X(x)}[b_i, b_j]^n \quad (31)$$

Because $b_i$ follows a Gaussian distribution, $\mathrm{E}_{P_X(x)}[\mathrm{d}\Theta(b_i)/\mathrm{d}b_i] = \mathrm{E}_{P_X(x)}[\delta(b_i)] = p_{b_i}(b_i = 0) = 1/\sqrt{2\pi\sigma_{b_i}^2}$ holds. Thus, from equation (31), we obtain $\mathrm{Cov}_{P_X(x)}[\Theta(b_i), \Theta(b_j)] = \mathrm{Cov}_{P_X(x)}[b_i, b_j]/(2\pi\sigma_{b_i}\sigma_{b_j}) = \mathrm{corr}_{P_X(x)}[b_i, b_j]/2\pi$ as the first-order approximation, where $\mathrm{corr}_{P_X(x)}[b_i, b_j]$ denotes the correlation between $b_i$ and $b_j$.

Hence, the variance of $c$ is computed as

$$\mathrm{Var}_{P_X(x)}[c] = \sum_{i=1}^{N}\sum_{j=1}^{N} \mathrm{Cov}_{P_X(x)}[\Theta(b_i), \Theta(b_j)] = \frac{N}{4} + \frac{1}{2\pi}\sum_{i=1}^{N}\sum_{j\neq i} \mathrm{corr}_{P_X(x)}[b_i, b_j] \quad (32)$$

For simplicity, we define a new effective variable (effective degree of $L(x)$) as

$$\widehat{K} := 1 + \frac{1}{N}\sum_{i=1}^{N}\sum_{j\neq i} \mathrm{corr}_{P_X(x)}[b_i, b_j] \quad (33)$$

Using this, we obtain $\lambda = 1 - \left(1 + \frac{2(\widehat{K}-1)}{\pi}\right)^{-1/2}$. Therefore, substituting $\lambda$ and $Z$ into equation (27), we obtain

$$N_{\mathrm{LM}} = P(c=0)2^N = \left(1 + \frac{2(\widehat{K}-1)}{\pi}\right)^{-\frac{1}{2}} \left(2 - \left(1 + \frac{2(\widehat{K}-1)}{\pi}\right)^{-\frac{1}{2}}\right)^N \quad (34)$$

Because $Z^{-1} = \left(1 + \frac{2(\widehat{K}-1)}{\pi}\right)^{-1/2}$ is of order one, we can omit it and obtain equation (3) as the order.

We refer to $\widehat{K}$ as the effective degree of $L(x)$ because, when the coefficients $\{a_i, a_{ij}, \dots, a_{i_1 \cdots i_K}\}$ are of the same order (referred to as $\sigma_a$), $\widehat{K}$ approximates the degree of $L(x)$, $\widehat{K} \approx K = \deg L(x)$. When the coefficients are of order $\sigma_a$, we can approximate $\mathrm{Cov}_{P_X(x)}[b_i, b_j]$ and $\sigma_{b_i}^2$ as $\mathrm{Cov}_{P_X(x)}[b_i, b_j] = a_{ij}^2 + \sum_{k \neq i, k \neq j} a_{ijk}^2 + \cdots \approx \sum_{\alpha=0}^{K-2} \binom{N-2}{\alpha} \sigma_a^2 \approx \binom{N-2}{K-2} \sigma_a^2$ and $\sigma_{b_i}^2 = \mathrm{Var}_{P_X(x)}[b_i] = a_i^2 + \sum_{j \neq i} a_{ij}^2 + \sum_{j \neq i, k \neq i, j<k} a_{ijk}^2 + \cdots \approx \sum_{\alpha=0}^{K-1} \binom{N-1}{\alpha} \sigma_a^2 \approx \binom{N-1}{K-1} \sigma_a^2$ as the leading order, which provides $\mathrm{corr}_{P_X(x)}[b_i, b_j] \approx \binom{N-2}{K-2} \sigma_a^2 / \binom{N-1}{K-1} \sigma_a^2 = \frac{K-1}{N-1}$. These approximations hold because $x_1, \ldots, x_N$ are independently sampled from $P_X(x)$. Therefore, from equation (33), we obtain $\widehat{K} \approx 1 + \frac{1}{N} \sum_{i=1}^{N} \sum_{j \neq i} \frac{K-1}{N-1} = K$ as the leading order.

**Derivation of Theorem 1**

The non-local search algorithm is defined as equation (4). Here, coefficient $f(x^{(m)})$ is the probability of $x_i^{(m)}$ being selected as the subsequent state $y_i$, where $0 \leq f(x^{(m)}) \leq 1$ and $\sum_{m=1}^{M} f(x^{(m)}) = 1$. The expectation of the subsequent state is given as $\mathbf{y} := \mathrm{E}_{P_Y(y)}[y] = \mathrm{E}_{f(x)}[x] = \sum_{m=1}^{M} f(x^{(m)}) x^{(m)}$, where $\mathrm{E}_{P_Y(y)}[\cdot] = \sum_{y \in Y} \cdot P_Y(y)$ indicates the expectation of $\cdot$ over $P_Y(y)$, and $\mathrm{E}_{f(x)}[\cdot] = \sum_{m=1}^{M} \cdot f(x^{(m)})$ indicates the expectation of $\cdot$ over $f(x)$.

Here, we extend equation (5) and define the objective function as follows:

$$\mathcal{L}(f(x)) = \mathrm{E}_{P_Y(y)}[L(y)] - \sqrt{2 \log M \, \mathrm{Var}_{P_Y(y)}[L(y)]} + \frac{1}{\beta} \mathrm{E}_{f(x)}[\log f(x)] \quad (35)$$

where a large positive constant $\beta$ indicates a hyperparameter (inverse temperature), and $-\mathrm{E}_{f(x)}[\log f(x)]$ indicates the Shannon entropy of $f(x)$. This serves to identify the optimal $f(x)$ that minimises $\mathcal{L}(f(x))$ under the constraints $0 \leq f(x^{(m)}) \leq 1$ and $\sum_{m=1}^{M} f(x^{(m)}) = 1$. When $\beta$ is sufficiently large, equation (35) converges to equation (5) because the last term of equation

(35) becomes negligibly smaller than the other terms. Thus, equations (5) and (35) are effectively the same objective function. The expectation of $L(y)$ over $P(y)$ is denoted as

$$L(\mathbf{y}) = \mathrm{E}_{P_Y(y)}[L(y)] = \mathrm{E}_{P_Y(y)}\left[\sum_{i=1}^{N} a_i y_i + \sum_{i<j} a_{ij} y_i y_j + \cdots\right] = \sum_{i=1}^{N} a_i \mathbf{y}_i + \sum_{i<j} a_{ij} \mathbf{y}_i \mathbf{y}_j + \cdots \quad (36)$$

or equivalently, $L(\mathbf{y}) = \mathrm{E}_{P_Y(y)}[\sum_{i=1}^{n} l(y)_i] = \sum_{i=1}^{n} l(\mathbf{y})_i$, because each element of $y$ is sampled in a mutually independent manner.

Now, we derive the approximates of $L(\mathbf{y})$ and $\mathrm{Var}_{P_Y(y)}[L(y)]$ to make $\mathcal{L}(f(x))$ analytically tractable. We approximate $L(\mathbf{y})$ using $\mathrm{E}_{f(x)}[L(x)] = \sum_{m=1}^{M} f(x^{(m)}) L(x^{(m)})$. To this end, we define a vector expression of $L(\mathbf{y})$, $l(\mathbf{y}) = (a_1 \mathbf{y}_1, a_2 \mathbf{y}_2, \ldots, a_{(N-K+1)\cdots N} \mathbf{y}_{N-K+1} \cdots \mathbf{y}_N)^\mathrm{T}$, which is an $n$-dimensional vector that arranges every component of $L(\mathbf{y})$ in each row. We assume that the $i$th component of $l(\mathbf{y})$, $l(\mathbf{y})_i$, is approximately proportional to the $i$th component of $l(x)$, $l(x)_i$, with a proportional coefficient $\rho$ when $x$ is sampled from $f(x)$. This means that $l(\mathbf{y})_i \approx \rho l(x)_i$ for $i = 1, \ldots, n$. Because $|l(x)|^2 = \sum_{i=1}^{n} l(x)_i^2 = \mathrm{Var}_{P_X(x)}[L(x)] = 1$ by construction, $|l(\mathbf{y})|^2 \approx \rho^2 |l(x)|^2 = \rho^2$ holds. Thus, by taking the average over $f(x)$, we obtain

$$L(\mathbf{y}) \approx |l(\mathbf{y})| \mathrm{E}_{f(x)}[L(x)] \quad (37)$$

This approximation is further justified in terms of the maximum likelihood estimation. Here, we suppose that $l(\mathbf{y})_i^2$ follows a Gaussian distribution $p(l(\mathbf{y})_i^2) = \mathcal{N}[\rho^2 l(x)_i^2, l(x)_i^2]$ and derive the maximum likelihood estimate of $\rho$ that minimises the following negative log likelihood:

$-\log p(l(\mathbf{y})_1^2, \ldots, l(\mathbf{y})_n^2) = -\sum_{i=1}^{n} \log p(l(\mathbf{y})_i^2) = \sum_{i=1}^{n} |l(\mathbf{y})_i^2 - \rho^2 l(x)_i^2|^2 / 2 l(x)_i^2 + \mathrm{const}$. The estimate is given as $\rho = \sqrt{|l(\mathbf{y})|^2} = |l(\mathbf{y})|$, which yields an approximation $L(\mathbf{y}) \approx |l(\mathbf{y})| L(x)$. This approximation also holds for the average of $L(x)$ over $f(x)$. Therefore, we obtain equation

(37).

Moreover, when the coefficients of $L(y)$ are sufficiently isotropic such that they satisfy inequality (7), following Lemma 1 (equation (13)), the variance of $L(y)$ is given as $\text{Var}_{P_Y(y)}[L(y)] = 1 - |l(\mathbf{y})|^2$ as the leading order for large N. Consequently, we obtain the following approximation of $\mathcal{L}(f(x))$:

$$\mathcal{L}(f(x)) \approx |l(\mathbf{y})|\mathbb{E}_{f(x)}[L(x)] - \sqrt{2\log M \,(1 - |l(\mathbf{y})|^2)} + \frac{1}{\beta}\mathbb{E}_{f(x)}[\log f(x)] \tag{38}$$

The optimal $f(x)$ is identified using a variational approach. The variation of $\mathcal{L}(f(x))$ with respect to $f(x)$ is given by $\delta\mathcal{L} = \int \left\{|l(\mathbf{y})|L(x) + \frac{1}{\beta}(\log f(x) + 1)\right\}\delta f(x)dx$. Thus, by solving $\delta\mathcal{L} = 0$, we find that the optimal $f(x)$ that minimises $\mathcal{L}(f(x))$ is a Boltzmann distribution in the following form:

$$f(x^{(m)}) \propto e^{-\beta|l(\mathbf{y})|L(x^{(m)})} \tag{39}$$

With a sufficiently large $\beta$, this distribution has a sharp peak at the state that minimises $L(x)$. For $x$ that minimises $L(x)$, $f(x)$ is close to 1; otherwise, $f(x)$ is close to 0. Thus, using a small positive constant $\gamma$ that satisfies $0 < \gamma \leq 0.5$, we assign the weights as $f(x^{(m_1)}) = 1 - \gamma$ for the best state that minimises the cost $m_1 = \underset{m}{\text{argmin}}\, L(x^{(m)})$, $f(x^{(m_2)}) = \gamma$ for the second-best state $m_2 = \underset{m \neq m_1}{\text{argmin}}\, L(x^{(m)})$, and $f(x^{(m)}) = 0$ otherwise.

Because $L(x^{(m_1)})$ and $L(x^{(m_2)})$ are the minimum and second minimum costs amongst M samples, $L(x^{(m_1)}) \approx L(x^{(m_2)}) \approx -\sqrt{2\log M}$ approximately holds for large M according to Lemma 2. Therefore, we obtain the proposed theorem (equation (6)).

**Notes on Theorem 1**

With the optimal $f(x)$, $\mathbf{y}$ becomes a function of $\gamma$, that is, $\mathbf{y} = (1-\gamma)x^{(m_1)} + \gamma x^{(m_2)}$, where $|l(\mathbf{y})| = 1$ for $\gamma = 0$ and $|l(\mathbf{y})| \ll 1$ for $\gamma = 0.5$.

The minimisation of the computational cost is achieved with the crossover of two states. This speaks to the emergence of the selection and crossover algorithm as a natural consequence of computational cost minimisation. However, in contrast to the usual selection and crossover in which the crossover weight is half and half, here we found that a biased crossover is crucial to maximise the search efficiency. The variance of the offspring created by selection and crossover is close to one unless the crossover rate $\gamma$ is notably small, enabling the offspring to exhibit a high diversity. In short, the balance of minimising the mean and maintaining a large variance enhances the search efficiency.

The selection and crossover algorithm identifies a solution with a cost lower than that of the random search, whereas their computational costs are the same. Remarkably, this scheme can be combined with gradient descent algorithms by replacing the original cost function $L(y)$ with the proxy cost function $L'(y)$, as described in the Results section. This means that the search efficiency of the optimal combination of the two algorithms with a computational cost of 2*M* (i.e., *M* for parents and *M* for offspring) is of the same order as the naive gradient descent algorithms with computational cost of $M^2$. This speaks to the quadratic speedup of the global search, providing a higher chance of identifying a better approximate global minimum within a limited timeframe. The proposed scheme exhibits the highest efficiency among possible search strategies based on combinations of past states considered in equation (4).

Because the costs for the top few parent states are all $-\sqrt{2\log M}$ as the leading order when $M$ is sufficiently large, a crossover of any pair among the top few parent states exhibits approximately the same efficiency. Although the combination of the gradient descent algorithm and the selection and crossover algorithm has a high search efficiency, it may fall into a local minimum. Thus, multiple sequences of the algorithm with different parents will increase the chance for determining a good approximate global minimum.

**Simulation protocols**

In the numerical simulations shown in **Figs. 1–3**, the optimisation of the $K$th-degree cost functions defined by equation (1) with $K$ = 2, 3, 4 was considered, where the coefficients of $L(x)$, $\{a_i, a_{ij}, \ldots, a_{i_1 \cdots i_K}\}$, were independently sampled from an identical Gaussian distribution prior to the optimisation and fixed during the optimisation.

For **Fig. 1c**, the global minimum was determined by sequentially searching all possible states for $N \leq 20$; otherwise, it was approximated by the best solution of gradient descent searches with $10^4$-times iterative random initialisation.

For **Fig. 2b**, we randomly sampled $M$ states and counted the number of states that satisfied the condition for a local minimum, that is, $L(x) < L(D_1 x), \ldots, L(x) < L(D_N x)$, where $M$ was scaled from $4 \times 10^6$ to $10^9$ depending on $N$ and $K$ because the probability of $x$ being a local minimum decreases as $N$ increases. Then, the number of local minima of the cost function $N_{LM}$ was computed by multiplying the obtained number by $2^N/M$.

For **Fig. 3**, the fourth-degree cost functions in a 200-dimensional binary state space ($N = 200$, $K = 4$) were employed. For **Fig. 3b–d**, $M = 2 \times 10^4$ parent states were randomly sampled from the parent distribution. The selection and crossover algorithm determined states providing the two minimum costs as the parents, and generated $M = 2 \times 10^4$ offspring states as a crossover of the parents' states, followed by the selection of the best offspring state with the minimum cost. In **Fig. 3c**, the theoretical predictions of the offspring's statistics were as follows: mean $\mu_Y = -\sqrt{2 \log M} \, |l(\mathbf{y})|$, standard deviation $\sigma_Y = \sqrt{1 - |l(\mathbf{y})|^2}$, and minimum among $M$ samples $L_{M,Y} = \mathcal{L}(f(x))$, which was obtained using equation (6). Conversely, the statistics of parents were $\mu_X = 0$, $\sigma_X = 1$, and $L_{M,X} = -\sqrt{2 \log M}$. However, to make a finer fit to the numerical results, $-\sqrt{2 \log M}$ in these theoretical values were replaced with their first-order perturbation obtained using equation (21).

For **Fig. 3e**, the gradient descent algorithm was conducted with $M = 100$ randomly selected initial states, the solutions of which give the parent distribution. Then, the selection and crossover algorithm determined the best state and one among the second-to-11th-best states as the parents, and generated $M = 100$ offspring states, followed by the gradient descent algorithm starting from the obtained 100 offspring states. This treatment of randomly selecting a parent from the second-to-11th-best states served to ensure that the offspring distribution has a sufficient diversity, because otherwise many offspring may converge to a local minimum that is the same as a parent state owing to the low effective state dimensionality of this experimental setup.

For **Fig. 4**, travelling salesman problems comprising $N = 500$ cities were considered. Costs were determined by the Euclidean distance on the plane. The cost function (i.e., total distance) was normalised prior to the optimisation to ensure zero mean and unit variance. For gradient descent search, the 2-opt algorithm [25] was conducted with *M* = 500 randomly selected initial routes. For selection and crossover, the 10 best routes that provide 10 minimum costs were stored, and two among them were randomly selected to generate *M* = 500 offspring states. A crossover rate of $\gamma = 0.05$ was used. This was followed by the 2-opt algorithm for the offspring.

**Data Availability**

All relevant data are presented within the paper. Figs. 1–4 were generated using the author's scripts (see Code Availability).

**Code Availability**

The MATLAB scripts are available at https://github.com/takuyaisomura/computational_cost.

**Acknowledgements**

T.I. is funded by RIKEN Center for Brain Science. The funder had no role in the study design, data collection and analysis, decision to publish, or preparation of the manuscript.

**Competing Interests**

The author has no competing interests to declare.

# References

1.  Wolpert, D. H. & Macready, W. G. No free lunch theorems for optimization. *IEEE Trans. Evol. Comput.* **1**(1), 67-82 (1997).

2.  Schumacher, C., Vose, M. D. & Whitley, L. D. The no free lunch and problem description length. *Proceedings of the Genetic and Evolutionary Computation Conference (GECCO-2001)* 565-570 (2001).

3.  Igel, C. & Toussaint, M. A no-free-lunch theorem for non-uniform distributions of target functions. *J. Math. Model. Algorithms* **3**(4), 313-322 (2005).

4.  Rowe, J. E., Vose, M. D. & Wright, A. H. Reinterpreting no free lunch. *Evol. Comput.* **17**(1), 117-129 (2009).

5.  Bishop, C. M. *Pattern Recognition and Machine Learning* (Springer, New York, 2006).

6.  Goodfellow, I., Bengio, Y. & Courville, A. *Deep Learning* (MIT Press, Cambridge, MA, 2016).

7.  Mitchell, M. *An Introduction to Genetic Algorithms* (MIT Press, Cambridge, MA, 1998).

8.  Sivanandam, S. N. & Deepa, S. N. *Introduction to Genetic Algorithms* (Springer, Berlin, Heidelberg, 2008).

9.  Weise, T. *Global Optimization Algorithms—Theory and Application* (Self-Published Thomas Weise, 2009).

10. Jansen, T., & Wegener, I. The analysis of evolutionary algorithms—A proof that crossover really can help. *Algorithmica* **34**(1), 47-66 (2002).

11. Kötzing, T., Sudholt, D. & Theile, M. How crossover helps in pseudo-Boolean optimization. *Proceedings of the 13th Annual Conference on Genetic and Evolutionary Computation* 989-996 (2011).

12. Dang, D. C. et al. Escaping local optima with diversity mechanisms and crossover. *Proceedings of the 2016 on Genetic and Evolutionary Computation Conference* 645-652 (2016).

13. Doerr, B., Happ, E. & Klein, C. Crossover can provably be useful in evolutionary computation. *Theor. Comput. Sci.* **425**, 17-33 (2012).

14. Katoch, S., Chauhan, S. S. & Kumar, V. A review on genetic algorithm: past, present, and future. *Multimed. Tools. Appl.* **80**(5), 8091-8126 (2021).

15. Miikkulainen, R. & Forrest, S. A biological perspective on evolutionary computation. *Nat. Mach. Intell.* **3**(1), 9-15 (2021).

16. Such, F. P. et al. Deep neuroevolution: Genetic algorithms are a competitive alternative for training deep neural networks for reinforcement learning. arXiv preprint arXiv:1712.06567 (2017).

17. Sehgal, A., La, H., Louis, S. & Nguyen, H. Deep reinforcement learning using genetic algorithm for parameter optimization. In *Third IEEE International Conference on Robotic Computing (IRC)* (pp. 596-601). IEEE (2019).

18. Stanley, K. O., Clune, J., Lehman, J. & Miikkulainen, R. Designing neural networks through neuroevolution. *Nat. Mach. Intell.* **1**(1), 24-35 (2019).

19. Iba, H. & Noman, N. *Deep Neural Evolution: Deep Learning with Evolutionary Computation.* (Springer, Singapore, 2020).

20. Korte, B. & Vygen, J. *Combinatorial Optimization Vol. 2* (Springer, Heidelberg, 2012).

21. Friston, K. J., Kilner, J. & Harrison, L. A free energy principle for the brain. *J. Physiol. Paris* **100**, 70-87 (2006).

22. Friston, K. J. The free-energy principle: a unified brain theory? *Nat. Rev. Neurosci.* **11**, 127-138 (2010).

23. Isomura, T. & Toyoizumi, T. On the achievability of blind source separation for high-dimensional nonlinear source mixtures. *Neural Comput.* **33**(6), 1433-1468 (2021).

24. Applegate, D. L., Bixby, R. E., Chvátal, V. & Cook, W. J. *The Traveling Salesman Problem: A Computational Study* (Princeton University Press, 2006).

25. Johnson, D. S. & McGeoch, L. A. The travelling salesman problem: A case study in local optimization. In E. H. L. Aarts & J. K. Lenstra (Eds.), *Local Search in Combinatorial Optimization* (pp. 215-310). John Wiley & Sons, Chichester, England (1997).

26. Jung-Beeman, M. et al. Neural activity when people solve verbal problems with insight. *PLoS Biol.* **2**(4), e97 (2004).

27. Bowden, E. M., Jung-Beeman, M., Fleck, J. & Kounios, J. New approaches to demystifying

insight. *Trends Cogn. Sci.* **9**(7), 322-328 (2005).

28. Kounios, J. & Beeman, M. The cognitive neuroscience of insight. *Annu. Rev. Psychol.* **65**, 71-93 (2014).

29. Tik, M. et al. Ultra-high-field fMRI insights on insight: Neural correlates of the Aha!-moment. *Hum. Brain Mapp.* **39**(8), 3241-3252 (2018).

30. Bengio, Y., Le Roux, N., Vincent, P., Delalleau, O. & Marcotte, P. Convex neural networks. *Adv. Neural Info. Proc. Syst.* **18**, 123 (2006).

31. Isomura, T. & Toyoizumi, T. Dimensionality reduction to maximize prediction generalization capability. *Nat. Mach. Intell.* **3**(5), 434-446 (2021).

32. Forrest, S. & Mitchell, M. Relative building-block fitness and the building-block hypothesis. In *Foundations of Genetic Algorithms* (Vol. 2, pp. 109-126), Elsevier (1993).

33. Stephens, C. & Waelbroeck, H. Schemata evolution and building blocks. *Evol. Comput.* **7**(2), 109-124 (1999).

34. Mnih, V. et al. Human-level control through deep reinforcement learning. *Nature* **518**(7540), 529-533 (2015).

35. Silver, D. et al. Mastering the game of Go with deep neural networks and tree search. *Nature* **529**(7587), 484-489 (2016).

36. Silver, D. et al. Mastering chess and shogi by self-play with a general reinforcement learning algorithm. arXiv preprint arXiv:1712.01815 (2017).

37. Isomura, T. & Friston, K. J. Reverse-engineering neural networks to characterize their cost functions. *Neural Comput.* **32**, 2085-2121 (2020).

38. Isomura, T., Shimazaki, H. & Friston, K. J. Canonical neural networks perform active inference. bioRxiv preprint:2020.12.10.420547 (2020).

39. Leleu, T., Yamamoto, Y., McMahon, P. L. & Aihara, K. Destabilization of local minima in analog spin systems by correction of amplitude heterogeneity. *Phys. Rev. Lett.* **122**(4), 040607 (2019).

40. Roth, G. & Dicke, U. Evolution of the brain and intelligence. *Trends Cogn. Sci.* **9**(5), 250-257 (2005).

41. Bullmore, E. & Sporns, O. The economy of brain network organization. *Nat. Rev. Neurosci.* **13**(5), 336-349 (2012).

42. Lumley, A. J. et al. Sexual selection protects against extinction. *Nature* **522**(7557), 470-473 (2015).

43. Friston, K. J. et al. Active inference, curiosity and insight. *Neural Comput.* **29**(10), 2633-2683 (2017).

44. Colin, T. R. & Belpaeme, T. Reinforcement learning and insight in the artificial pigeon. In *41st Annual Meeting of the Cognitive Science Society (CogSci 2019)* (pp. 1533-1539). Cognitive Science Society (2019).

45. Linsker, R. Self-organization in a perceptual network. *Computer* **21**, 105-117 (1988).

46. Dayan, P., Hinton, G. E., Neal, R. M. & Zemel, R. S. The Helmholtz machine. *Neural Comput.* **7**,

889-904 (1995).

47. Sutton, R. S. & Barto, A. G. *Reinforcement Learning* (MIT Press, Cambridge, MA, 1998).

48. Sussillo, D. & Abbott, L. F. Generating coherent patterns of activity from chaotic neural networks. *Neuron* **63**, 544-557 (2009).

49. Laje, R. & Buonomano, D. V. Robust timing and motor patterns by taming chaos in recurrent neural networks. *Nat. Neurosci.* **16**, 925-933 (2013).

50. Wolpert, D. M. & Kawato, M. Multiple paired forward and inverse models for motor control. *Neural Netw.* **11**(7-8), 1317-1329 (1998).

51. Isomura, T., Parr, T. & Friston, K. J. Bayesian filtering with multiple internal models: toward a theory of social intelligence. *Neural Comput.* **31**(12), 2390-2431 (2019).

52. Hayashi-Takagi, A. et al. Labelling and optical erasure of synaptic memory traces in the motor cortex. *Nature* **525**(7569), 333-338 (2015).

53. Holtmaat, A. & Caroni, P. Functional and structural underpinnings of neuronal assembly formation in learning. *Nat. Neurosci.* **19**(12), 1553-1562 (2016).

54. Merolla, P. A. et al. A million spiking-neuron integrated circuit with a scalable communication network and interface. *Science* **345**(6197), 668-673 (2014).

55. Roy, K., Jaiswal, A. & Panda, P. Towards spike-based machine intelligence with neuromorphic computing. *Nature* **575**(7784), 607-617 (2019).

56. Grover, L. K. A fast quantum mechanical algorithm for database search. *Proceedings of the*

*28th Annual ACM Symposium on Theory of Computing* 212-219 (1996).

57. Bennett, C. H., Bernstein, E., Brassard, G. & Vazirani, U. Strengths and weaknesses of quantum computing. *SIAM J. Comput.* **26**(5), 1510-1523 (1997).

58. Toyoizumi, T. & Abbott, L. F. Beyond the edge of chaos: Amplification and temporal integration by recurrent networks in the chaotic regime. *Phys. Rev. E.* **84**(5), 051908 (2011).

**Table 1. Glossary of expressions**

| Expression | Description |
|:---:|:---:|
| $N$ | State dimensionality |
| $M$ | Number of samples |
| $X = \{-1, 1\}^N$ | Set of parent states |
| $Y \subseteq X$ | Set of offspring states |
| $x = (x_1, \ldots, x_N)^T \in X$ | Parent state |
| $y = (y_1, \ldots, y_N)^T \in Y$ | Offspring state |
| $P_X(x) = \prod_{i=1}^{N} P_X(x_i)$ | Parent distribution |
| $P_Y(y) = \prod_{i=1}^{N} P_{Y_i}(y_i)$ | Offspring distribution |
| $L(x)$ | Parent cost |
| $L(y)$ | Offspring cost |
| $K = \dim L(x)$ | Degree of $L(x)$ |
| $a_{i_1 \cdots i_\alpha}$ | Coefficient of $L(x)$ |
| $L_M$ | Minimum cost among $M$ samples |
| $N_{\text{LM}}$ | Number of local minima of $L(x)$ |
| $\widehat{K}$ | Effective degree of $L(x)$ |
| $L'(x)$ | Proxy cost function |
| $f(x)$ | Weight distribution |
| $\gamma$ | Crossover rate |
| $l(\mathbf{y})$ | Vector expression of $\mathrm{E}_{P_Y(y)}[L(y)]$ |
| $\|l(\mathbf{y})\|$ | Norm of $l(\mathbf{y})$ |
| $\mu_Y = \mathrm{E}_{P_Y(y)}[L(y)]$ | Expectation of $L(y)$ over $P_Y(y)$ |
| $\sigma_Y^2 = \mathrm{Var}_{P_Y(y)}[L(y)]$ | Variance of $L(y)$ over $P_Y(y)$ |